\documentclass[sn-apa, iicol]{sn-jnl}

\usepackage{graphicx}%
\usepackage{multirow}%
\usepackage{amsmath,amssymb,amsfonts}%
\usepackage{amsthm}%
\usepackage{mathrsfs}%
\usepackage[title]{appendix}%
\usepackage{textcomp}%
\usepackage{manyfoot}%
\usepackage{booktabs}%
\usepackage{algorithm}%
\usepackage{algorithmicx}%
\usepackage{algpseudocode}%
\usepackage{listings}%
\usepackage[table]{xcolor}
\usepackage{appendix}
\usepackage{float}
\usepackage{subcaption}
\usepackage{caption}
\usepackage{siunitx}
\usepackage{rotating}
\usepackage{lscape}
\usepackage{verbatim}
\usepackage{titletoc}
\usepackage{alphalph}
\usepackage{tabularx}
\usepackage{adjustbox}
\usepackage{natbib}

\usepackage[colorinlistoftodos]{todonotes}

\usepackage[trackchanges,commandnameprefix=ifneeded]{changes}
\definechangesauthor[name={Shijia Feng}, color=red]{SF}

\theoremstyle{thmstyleone}%
\theoremstyle{thmstyletwo}%

\theoremstyle{thmstylethree}%

\begin{document}

\title[Article Title]{Are you Struggling? Dataset and Baselines for Struggle Determination in Assembly Videos}


\author[1]{\fnm{Shijia} \sur{Feng}}\email{shijia.feng.2019@bristol.ac.uk}

\author[1]{\fnm{Michael} \sur{Wray}}\email{michael.wray@bristol.ac.uk}

\author[2]{\fnm{Brian} \sur{Sullivan}}\email{brian.sullivan@bristol.ac.uk}

\author[5]{\fnm{Youngkyoon} \sur{Jang}}\email{youngkyoonjang@gmail.com}

\author[3]{\fnm{Casimir} \sur{Ludwig}}\email{c.ludwig@bristol.ac.uk}

\author[3]{\fnm{Iain} \sur{Gilchrist}}\email{i.d.gilchrist@bristol.ac.uk}

\author*[1,4]{\fnm{Walterio} \sur{Mayol-Cuevas}}\email{walterio.mayol-cuevas@bristol.ac.uk}

\affil*[1]{\orgdiv{School of Computer Science}, \orgname{University of Bristol}, \orgaddress{\city{Bristol}, \country{UK}}}

\affil[2]{\orgdiv{Bristol Medical School (PHS)}, \orgname{University of Bristol}, \orgaddress{\city{Bristol}, \country{UK}}}

\affil[3]{\orgdiv{School of Psychological Science}, \orgname{University of Bristol}, \orgaddress{\city{Bristol}, \country{UK}}}

\affil[4]{\orgname{Amazon}, \orgaddress{\city{Seattle}, \state{WA}, \country{United States}}}

\affil[5]{\orgdiv{Work done while at the} \orgname{University of Bristol}}


\abstract{Determining when people are struggling allows for a finer-grained understanding of actions that complements conventional action classification and error detection. Struggle detection, as defined in this paper, is a distinct and important task that can be identified without explicit step or activity knowledge. We introduce the first struggle dataset with three real-world problem-solving activities that are labelled by both expert and crowd-source annotators. Video segments were scored w.r.t. their level of struggle using a forced choice 4-point scale. This dataset contains 5.1 hours of video from 73 participants. We conducted a series of experiments to identify the most suitable modelling approaches for struggle determination. Additionally, we compared various deep learning models, establishing baseline results for struggle classification, struggle regression, and struggle label distribution learning. Our results indicate that struggle detection in video can achieve up to $88.24\%$ accuracy in binary classification, while detecting the level of struggle in a four-way classification setting performs lower, with an overall accuracy of $52.45\%$. Our work is motivated toward a more comprehensive understanding of action in video and potentially the improvement of assistive systems that analyse struggle and can better support users during manual activities.}


\maketitle

\section{Introduction}
\label{sec:intro}

\begin{figure*}[ht]
  \centering
  \includegraphics[width=0.9\linewidth]{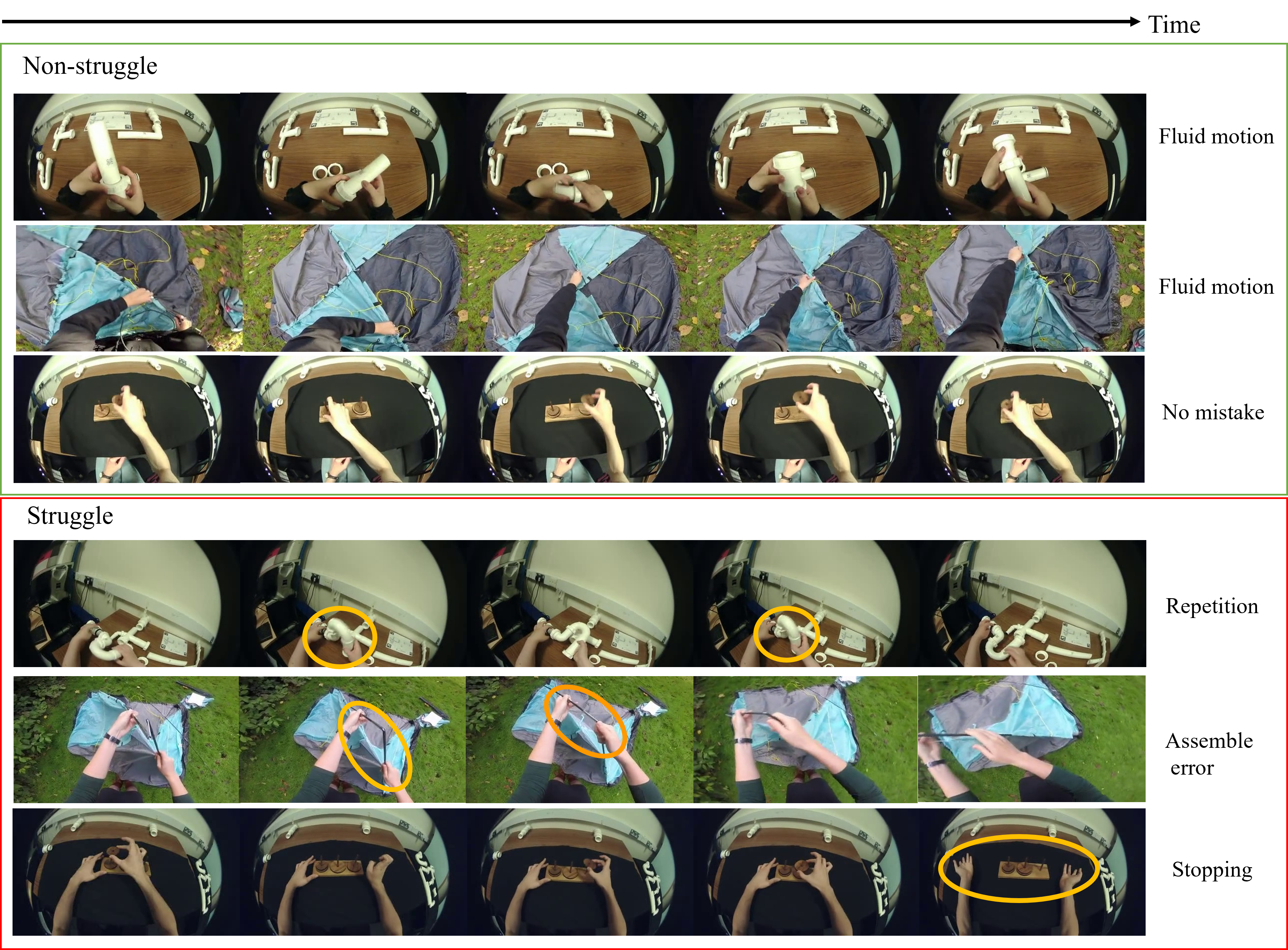}
  \caption{\textbf{Struggle and non-struggle examples.} \textcolor{green}{upper box} Three non-struggle examples from Pipes-Struggle, Tent-Struggle, and Tower-Struggle. Three struggle examples from the same activities are shown in the \textcolor{red}{bottom box}. Frames are displayed in chronological order from left to right. Non-struggle examples feature fluid motion without errors, whereas videos showcasing struggling show issues (highlighted in orange).
  For example: repeatedly placing the item back (\textcolor{red}{top row struggle}), assembling errors (\textcolor{red}{middle row struggle}), and hesitation towards the next step (\textcolor{red}{bottom row struggle}).}
  \label{fig:struggling_illustrations}
\end{figure*}

\begin{figure*}[ht]
  \centering
  \includegraphics[width=0.9\linewidth]{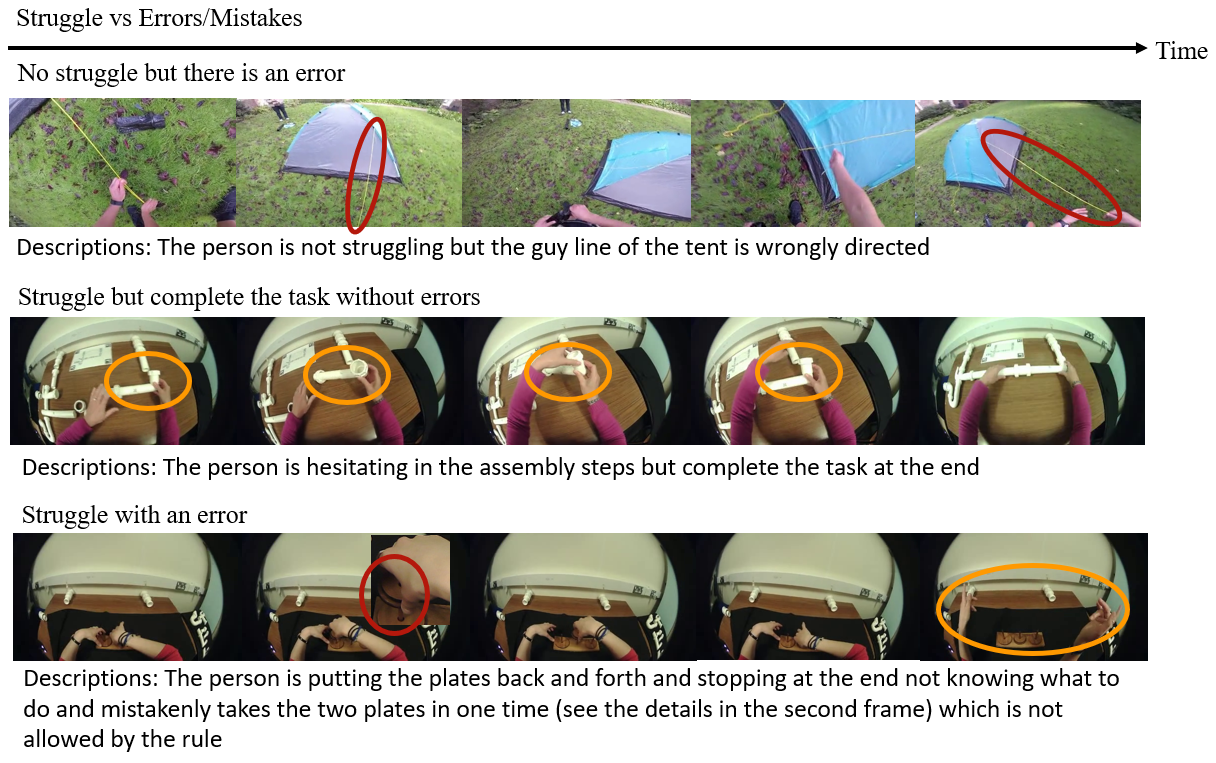}
  \caption{\textbf{Struggle vs Errors/Mistakes.} The figure showcases the differences between error detection and struggle determination using video examples from our dataset: (1) participant showing no sign of struggle, but there was an error in the completed activity; (2) participant showing signs of struggle but completed the activity successfully without any errors; and (3) participant showing signs of struggle with an error in the completed activity.}
  \label{fig:struggling_vs_error}
\end{figure*}

The ability to identify when someone is struggling is an important cognitive competence and a key component of skill acquisition \citep{Newell:etal:1991}. Observers can identify instances of struggle even when lacking specific task knowledge (e.g. a non-rock climber can determine when expert rock climbers are struggling). The ability to detect struggle and subsequently offer help has also been demonstrated in pre-verbal children and even other primates \citep{Warneken1301}. These findings suggest that struggle determination is a rather important and fundamental ability of visual action understanding which we define as a task within this paper---see examples within Figure~\ref{fig:struggling_illustrations}. 
However, most research to date in skill assessment has focused on error detection \citep{Ghoddoosian_2023_ICCV, HoloAssist2023, sener2022assembly101largescalemultiviewvideo, grauman2024egoexo4dunderstandingskilledhuman, Flaborea_2024_CVPR}, ranking of skill \citep{Doughty_2017_CVPR,Doughty_2019_CVPR} or task scoring \citep{https://doi.org/10.48550/arxiv.1611.05125, pirsiavash2014assessing}. Struggle, however, is distinct from error detection and these other tasks. It is possible to complete activities without step mistakes while struggling or to confidently complete an activity full of step mistakes without showing signs of struggle. Figure~\ref{fig:struggling_vs_error} presents examples from our dataset illustrating struggle-error relations. Struggle and error detection, though related, are distinct markers of action assessment. Importantly, the struggle is detectable without prior task knowledge, can be used to anticipate errors, can serve as an online scoring predictor before tasks are completed, and can assist an assistive system in identifying common steps that are difficult to perform. 
Therefore, we argue that visual understanding systems should incorporate both error and struggle detection. Struggle, a subjective behavioural marker, is challenging to define precisely. Nevertheless, it is easily detectable by non-experts, allowing for the construction of relevant datasets. In this work, we define struggle through the observations of annotators, where we observe that instances of struggle are identified during periods of motor hesitation (e.g. stopping or showing placement indecision), getting stuck or not knowing how to proceed, actions taking longer than expected; having non-smooth hand or body motions; frequent pauses and repeated attempts; or showing signs of frustration (e.g. through hand and or head movements).

Motivated by the importance of struggle determination within the larger goal of finer-grained visual understanding and the potential opportunities for eye-wear computing to provide real-time support assistance, we collected a dataset for struggle determination using indoor and outdoor real-world activities, including three domains: assembling plumbing pipes; pitching a camping tent; and solving the `Tower of Hanoi' puzzle. These three domains were chosen specifically to demonstrate the viability of struggle detection due to their task structures and assembly requirements. The Tower of Hanoi puzzle presented a strict rule-based framework requiring participants to solve a spatial puzzle. Pipe plumbing uses rigid objects and printed instructions for participants to follow, showcasing their ability to perform an unknown task with guidance. The tent assembly task introduced a free-form assembly process with deformable objects, demanding greater improvisation and manual skill. Our chosen tasks are examples of real-world problem-solving that require structured, sequential assembly with a predefined final state. These tasks involve following specific rules and constraints to achieve a well-defined outcome, such as correctly connecting components or completing a structured puzzle. Our tasks allow participants to demonstrate various complex visuomotor behaviours with multiple, albeit constrained, opportunities for struggle.

Our struggle-determination approach is based on two main assumptions. First, we believe there is sufficient information for many activities to determine if a person is struggling from video, based on low-level motion cues of hand-object interaction and without requiring specific task knowledge. This is supported by behavioural evidence, e.g. \cite{Warneken1301,Warneken:online} in people. 
Second, the same deep learning models and related training and testing strategies can be used for different tasks as they share common visual cues that help to determine struggle, such as the smoothness of hand motion.

Figure \ref{fig:struggling_illustrations} shows sample video frames from our data. We used two levels for annotation: MTurk workers and a domain expert. Annotations used a forced choice, four-level struggle scale so that we can determine if a person is struggling and to what degree. We evaluated several video understanding methods trained on our dataset and provided various baseline results across our proposed struggle determination activities and machine learning tasks.  

In summary, our contributions are three-fold: 
\begin{enumerate}
\item We define the problem of struggle determination as a fine-grained task from egocentric assembly videos and introduce a public dataset for struggle determination. This includes new data subsets for activities Pipes-Struggle and Tower-Struggle, as well as struggle annotation for activity Tent-Struggle from EPIC-Tent \citep{9022634}. All data is annotated using both an expert annotator and crowd-sourcing.
\item We conducted experiments to identify the most suitable modelling approach for struggle determination and to compare the performance of different deep models to determine the most efficient one. These results serve as baseline benchmarks for struggle determination, not only as a classification task but also through regression and label distribution learning as alternative approaches.
\item We demonstrate the essential role of motion information in struggle determination and present preliminary results on the generalization across the three activities.
\end{enumerate}

\section{Related Work}
\label{sec: related work}

We first review related prior video understanding research, highlighting action recognition, action quality assessment, skill ranking and video anomaly detection. We also summarize related datasets and make comparisons to our struggle determination offering, see Table \ref{tab:related-datasets} in the Appendix. 

{\bf Action Recognition.} Action recognition (AR) aims to classify actions from video. Many mainstream deep architectures are initially designed for AR but have been used for other video tasks such as Action Segmentation/Detection \citep{kalogeiton2017action, Chen_2021_ICCV, Zhao_2022_CVPR}, Action Anticipation \citep{Ego4D2022CVPR}, and video-text retrieval \citep{liu2019use,gabeur2020multi}.
Current AR datasets differ in granularity between coarse-grained actions, in which the background context displays large differences between classes \citep{soomro2012ucf101,Kuehne11,https://doi.org/10.48550/arxiv.1705.06950,caba2015activitynet}, and fine-grained actions in which models need to discriminate between action-motion information such as hand-object interactions \citep{Damen2018EPICKITCHENS,Zhang_2021_CVPR,https://doi.org/10.48550/arxiv.1706.04261,li2018resound,shao2020finegym,https://doi.org/10.48550/arxiv.2204.03646}.

Numerous deep architectures have been developed for action recognition (AR). Initially, 2D-ConvNets were designed to distinguish video actions from 2D spatial features in video frames. These models typically used temporal average pooling and/or optical flow to enhance motion understanding. 
Subsequently, 3D-ConvNets \citep{tran2015learning, carreira2017quo, DBLP:journals/corr/abs-2004-04730, xie2018rethinking, tran2019video, tran2018closer, feichtenhofer2019slowfast} gained popularity due to their ability to extract spatial-temporal features through 3D convolution operations. Benefiting from large-scale video datasets such as Kinetics \citep{https://doi.org/10.48550/arxiv.1705.06950}, 3D-ConvNets, despite having a larger number of parameters to train, can be trained end-to-end on these extensive datasets and transferred to numerous downstream tasks. Among the commonly used 3D-ConvNets, the I3D \citep{carreira2017quo} is notable for its use of RGB frames and optical flow as dual inputs to improve the extraction of motion features. On the other hand, the SlowFast Networks \citep{feichtenhofer2019slowfast} achieve state-of-the-art results within the realm of 3D-ConvNets across many benchmark datasets. These networks require only RGB frames as input, with the Fast pathway extracting motion features that are then fused with the Slow pathway, thus saving considerable time by eliminating the need for optical flow calculations.

Recently, Transformer-based architectures such as Vision Transformer (ViT) \citep{https://doi.org/10.48550/arxiv.2010.11929}, Swin Transformer \citep{liu2021swin}, MLP-Mixer \citep{DBLP:journals/corr/abs-2105-01601}, and gMLP \citep{DBLP:journals/corr/abs-2105-08050} have been known to achieve the state-of-the-art results on many visual tasks benefited with the attention layers and MLP layers which are also data-hungry. For AR, those architectures originally designed for handling images are adapted for handling the spatial-temporal dependencies in video data, including hierarchical architecture such as ViViT \citep{https://doi.org/10.48550/arxiv.2103.15691} and Video Transformer Networks (VTN) \citep{Neimark_2021_ICCV}, and the integration of spatial and temporal attention modules such as TimeSformer \citep{DBLP:journals/corr/abs-2102-05095}, MViTv1 \citep{Fan_2021_ICCV}, and MViTv2 \citep{li2022mvitv2}. Finally, the Visual-Language Model (VLM) has brought attention to the research field where people are trying to train deep models for more general question-answering purposes, including querying the video data, such as Video-LLaVA~\citep{lin2023video}. 

{\bf Action Quality Assessment.} Action quality assessment (AQA) is a regression task predicting action quality scores. Most datasets, such as AQA-7 \citep{parmar2019action}, Olympic Scoring Dataset \citep{https://doi.org/10.48550/arxiv.1611.05125, pirsiavash2014assessing}, JIGSAW \citep{gao2014jhu}, and MTL-AQA \citep{mtlaqa}, contain sports scenes, where annotations represent the scores given by expert human judges. The typical architecture for AQA consists of 3D convolutional neural networks (C3D) \citep{tran2015learning} to extract spatial-temporal features, followed by average pooling (C3D-AVG), LSTM \citep{HochSchm97} (C3D-LSTM), or attention modules to further aggregate information from various video segments and generate regression or multitask predictions \citep{mtlaqa}.
Some methods tackle this task using Label Distribution Learning~\citep{tang2020uncertainty} in which the distribution of assessment scores is predicted. Methods often predict the scores directly using MLPs~\citep{zhang2021auto} or the mean/std. dev of a Gaussian Distribution based on Variational Auto-Encoders (VAE)~\citep{tang2020uncertainty,https://doi.org/10.48550/arxiv.2207.14513}.

{\bf Skill Determination.} Similar to AQA, Skill Determination~\citep{10.1007/978-3-030-34255-5_24, Ogul2022, Doughty_2017_CVPR, Doughty_2019_CVPR, Whodance2023, li2019manipulation} aims to predict the skill level of the action being performed but does so via ranking. \cite{10.1007/978-3-030-34255-5_24, Ogul2022} mainly focus on ranking surgery skills using Siamese LSTM-based networks by aggregating kinematic data. \cite{Doughty_2017_CVPR} investigated ranking skills for general daily activities and introduced the EPIC-Skill2018 dataset. \cite{li2019manipulation} proposes an RNN-based spatial attention model and significantly improves the results in \cite{Doughty_2017_CVPR}.
\cite{Doughty_2019_CVPR} employs a combination of contrastive learning and attention to discriminate between high- and low-skill actions that require fine-grained discrimination skills. Most recently, \cite{Whodance2023} proposes a TikTok dance video dataset for ranking dancing performances. Ego-Exo4D \citep{grauman2024egoexo4dunderstandingskilledhuman} is a large-scale egocentric dataset that has been released concurrent to this research, with a proficiency estimation benchmark to infer the skill level of participants from six scenarios.

{\bf Video Anomaly Action Detection.} Video Anomaly Detection aims to detect unintentional~\citep{https://doi.org/10.48550/arxiv.2209.11870, zatsarynna2022selfsupervised} or anomalous actions~\citep{https://doi.org/10.48550/arxiv.2208.11113, sultani2018real, tian2021weaklysupervised, DBLP:journals/corr/abs-1907-10211}.
Whilst fine-grained discrimination may share some similarities to Struggle Detection, we differ in two major aspects: persons struggling may not contain anomalies, and Anomaly Detection is constructed as a temporal localisation task.

{\bf Error/Mistakes Detection.}
Several datasets in error/mistake detection are also conceptually related to and complementary to struggle. For example, Assembly101 \citep{sener2022assembly101largescalemultiviewvideo} is known as a large-scale dataset for procedure video understanding and is also one of the first datasets to include the novel task of detecting mistakes in procedural activities. In this dataset, participants are asked to assemble LEGO toy cars while being recorded by eight static cameras capturing the process from surrounding third-person viewpoints and four egocentric cameras capturing from first-person viewpoints. The mistake actions primarily involve procedural mistakes, such as putting on a part too early and making it impossible to assemble subsequent parts, which usually require prior knowledge to determine the correct sequence. There are also skill levels of the participants annotated, ranging from 1 (worst) to 5 (best). Besides, Anomalous Toy Assembly \citep{Ghoddoosian_2023_ICCV} contains toy assembly videos recorded from four different viewpoints and is annotated for when the participants display sequential anomalies in the assembly task, which mainly focuses on the procedural error detection. Holoassist \citep{HoloAssist2023} contains multi-modality video recordings in instructor-performer pairs for training the interactive AI assistant systems. During the task performance, the instructor may capture the mistakes of the performer and intervene in the task completion procedure, including helping the performer correct the mistakes. These datasets for error/mistake detection mainly focus on procedural mistakes, which are mostly related to understanding the right sequence of steps to perform a task and usually require domain prior knowledge to be annotated. Most of the methods developed to detect procedural mistakes are similar to anomaly detection pipelines. For example, \cite{Flaborea_2024_CVPR} proposed PREGO, in which the model is trained to anticipate the next steps from normal videos without mistakes and detect mistakes by calculating the distance from the feature of the actual videos during the test stage to see how much difference there are. This PREGO model is recognized as the first online one-class classification model for mistake detection in procedural egocentric videos and comes with two datasets adapted from Assembly101 \citep{sener2022assembly101largescalemultiviewvideo} and EPIC-Tent \citep{9022634}. 
In contrast, our struggle determination, as introduced in Section \ref{sec:intro}, though correlated with some motor errors, is more general and can be determined without prior knowledge, and in many cases, is a complement to the mistakes/errors. 

We find that fine-grained video understanding has been explored in a variety of different datasets and tasks (see Table \ref{tab:related-datasets} in the Appendix).
However, despite being crucial for fine-grained activity and skill understanding, determining the participant's level of struggle has not been attempted.
Previous research efforts motivate us to explore three modelling approaches for our dataset: struggle classification, struggle level regression, and struggle label distribution learning.
As labels for these tasks do not exist, we annotate struggle for one existing dataset~\citep{9022634, sullivan2021look} and collect and annotate two new activities.
Our three activities were chosen to require complex visuomotor coordination and problem-solving while remaining goal-driven. This promotes similar yet varied occurrences of struggle behaviours and provides enough diversity within and across activities.

\section{Struggle Determination Dataset}
\label{sec:dataset}

Here we detail the methodology to record our egocentric videos and collect annotations for the struggle determination dataset. Our struggle dataset is presented under the Non-Commercial Government Licence for public sector information \citep{nationalarchivesCommercialGovernment}. 

We initially recorded the participants' behaviour in three scenarios that require complex visuomotor coordination and problem-solving and in which participants may occasionally struggle to complete the tasks. 
These scenarios were intended to be similar to what one might encounter in daily life (both indoors and outdoors) and where one might want to receive supporting advice.
Finding suitable scenarios for struggle detection presents a balancing act. The scenarios shouldn't be too easy, as participants might not exhibit struggle behaviours, but they also shouldn't be overly complex or open-ended to prevent unclear goals and excessive task completion times (we favour activities contained within a few minutes). 

To effectively study struggle, an ideal video dataset should meet the following criteria: (1) The videos should be recorded with unobstructed views (e.g., egocentric) of a person performing a skill; (2) The dataset should contain videos of people performing a skill repeatedly; (3) The skills depicted in the video recordings should be such that they are likely to show instances of struggle.

We reviewed several existing datasets and found that most do not fully meet these criteria, particularly the second and third. Table~\ref{tab:related-datasets} in the Appendix compares these datasets in detail.

For example, EPIC-Kitchens~\citep{Damen2018EPICKITCHENS} consists of egocentric videos, satisfying the first criterion. However, the dataset primarily features daily cooking activities, which are not consistently repeated across participants, making it less suitable for studying struggle. Similarly, Ego4D Goal-Step~\citep{NEURIPS2023_7a65606f} mainly captures cooking tasks, but the videos vary by recipe, making it difficult to find repeated instances of the same task with both struggling and non-struggling cases.

Some prior datasets do contain repeated activities that could be suitable for struggle annotation. For example, \cite{9022634, sullivan2021look} features multiple subjects repeatedly assembling a camping tent—a complex visuomotor task where struggle is more evident. We annotate this data in our work.

Other datasets, such as MECCANO~\citep{Ragusa_2021_WACV, ragusa_MECCANO_2023} and Assembly101~\citep{sener2022assembly101largescalemultiviewvideo}, contain videos of participants assembling LEGO toys. Assembly101 also includes annotations for procedural mistakes and skill levels. However, neither dataset explicitly annotates struggle, and their focus on LEGO assembly makes them narrower in scope for our study.

Therefore, although many existing datasets exist, we decided to collect our own because our research focuses specifically on user struggle, which is not well-represented in those datasets. 
More importantly, we aimed for scenarios that, while diverse, would be ordinary enough for naive observers to estimate whether a participant was struggling without prior knowledge. To achieve this, we recorded first-person videos of two new activities, assembling plumbing pipes and game of Hanoi tower, and collected the struggle annotations of these videos from human observers. Additionally, we included the existing EPIC-Tent Dataset~\citep{9022634, sullivan2021look}, which also captures a relevant struggle-prone task of pitching a camping tent. 
Our dataset includes three scenarios: assembling plastic plumbing pipes by following one of two diagrams (easy or difficult) while seated indoors (Pipes), solving a version of the Tower of Hanoi puzzle while seated indoors (Tower), and assembling a camping tent while moving freely outdoors (Tent). The corresponding subsets for struggle determination are named Pipes-Struggle, Tower-Struggle, and Tent-Struggle. 
Each scenario consists of fine-grained actions that people perform routinely (e.g., arranging, picking up, and placing objects, attaching them, etc.). However, they differ in required equipment, difficulty, and goals. Participant performance also varies, with some struggling more than others when completing a scenario.

\begin{table*}[ht]
\caption{\textbf{Struggle level descriptions.}}
\label{tab:struggle-levels}
\resizebox{\textwidth}{!}{%
\begin{tabular}{l|l}
\toprule
\textbf{Definitely non-struggling} &
  \parbox{\textwidth}{The actions are executed with complete ease and confidence, with no indication of any challenge or difficulty encountered throughout the task.} \\ \midrule
\textbf{Slightly non-struggling} &
  \parbox{\textwidth}{The actions are generally executed smoothly and with confidence, with only occasional minor indications for some level of challenge or difficulty.} \\ \midrule
\textbf{Slightly struggling} &
  \parbox{\textwidth}{There are some signs of challenge or difficulty, including occasional errors, pauses, dropping off items, or wobbly hands that indicate the participants may not be very experienced in the task at hand but are still making an effort to complete it.} \\ \midrule
\textbf{Definitely struggling} &
  \parbox{\textwidth}{There are clear, consistent signs of difficulty, such as repeated attempts, unsmooth motions, and significant pauses during the task. These actions indicate that the participants don't really know what to do and may lack the knowledge or skill necessary to complete the task successfully.} \\ 
  \bottomrule
\end{tabular}%
}
\end{table*}
\subsection{Dataset Construction} 
\label{sec:dataset_construction}

\subsubsection{Data Collection}
For all three scenarios, university undergraduate and graduate student populations were recruited. All participants gave informed consent and approval for data sharing. We recruited 24 individuals to perform the plumbing task (80\% female, mean age 22.6 years), and 20 participants from this group also performed the Tower of Hanoi task. For Struggle-Tent, as stated in \cite{9022634}, 29 individuals (71\% female, mean age 23.3 years) performed the tent-pitching scenario. In all activities, participants wore a head-mounted GoPro Hero 5 (1080p at 60Hz). Videos were captured at $1920\times1080$ resolution, at 30Hz for Tent-Struggle, and 60Hz for Pipes-Struggle and Tower-Struggle.  

In the plumbing task, participants assembled home plumbing pipes following a diagram of a double sink layout (see supplementary material). Participants were evenly divided between those given an easier diagram and those given a more difficult one. To simplify dexterity requirements, the activity was reduced to assembling the layout of the pipes on a table rather than in a real sink. For Struggle-Tent, we focused on video segments from five actions in EPIC-Tent \citep{9022634}: assemble support, insert stake, insert support, insert support tab, and place guyline. Our expert annotator identified these actions as having the most obvious visual cues of struggle based on a series of pilot studies.
 
The video data was segmented into 10-second clips, and the struggle determination annotation was accomplished by one expert and MTurkers, both of whom rated these short video clips from the three activities. We choose to annotate struggle using video-level weak labels based on the 10-second windows for the following reasons: 
(1) Video-level weak labels simplify the annotation process, making it easier for annotators and easier to scale. During early annotation experiments, we observed that the annotators expressed difficulty locating the beginning and end of struggle segments accurately. 
(2) Annotating video-level weak labels has been found to be much more time-efficient than annotating the start and end frames in long, untrimmed videos (e.g. six times more time efficient~\cite{ma2020sfnetsingleframesupervisiontemporal}. When annotators mark the start and end frames of temporal segments, they often need to replay videos multiple times to verify boundaries. In contrast, annotating small video segments with a single segment-level label for struggle reduces the need for repeated viewings.
(3) Video-level weak labels help mitigate issues with the disagreement of the temporal action boundaries among the annotators, especially for more conceptual annotations, such as struggle. This was found in previous work~\cite{moltisanti2017trespassing}, in which researchers uncovered inconsistencies in the ground truth temporal boundaries across annotators.
By focusing on whether a segment contains struggle, ambiguity is reduced to determining the presence of struggle within the segment.
(4) We choose 10 seconds for the size of the annotation window following evidence from Experimental Psychology research that shows that manipulation struggle actions can be identified well within 10 seconds~\citep{Warneken1301, Warneken:online}. 

Based on the arguments above, our annotation clips thus were created by uniformly splitting the videos into 10-second segments from beginning to end (e.g. 0-9s, 10-19s, etc.). We also resized videos to $456\times256$ at 30Hz. This ensured segments had a sufficient resolution for both human observers and the deep learning models. Our annotation process thus focuses on evaluating struggle within isolated 10-second clips. This means that we assess and label the presence of struggle solely within each clip without considering the broader context of the full video recording.

\subsubsection{Annotating Struggle}

Since the original videos are uniformly trimmed into 10-second segments, we focus on `low-level' struggling cues, which we believe are the most transferable between participants and scenarios. Signs of struggle may include shaky hands, pausing actions, dropping items, or action repetition, however, we note that these are often task dependent.
While there is a variety of ways one might ask an observer to estimate struggle, we chose to ask observers to rate video segments using a four-point scale varying from definitely non-struggle to definitely struggle. Table \ref{tab:struggle-levels} lists the descriptions for each struggle level.
The value of a finer, four-point gradation of struggle supports the longer-term goal of assistive systems. For example, the support provided to a person who is `slightly' struggling may consist of a few basic instructions or reminders. However, a person determined to be definitely struggling may require more detailed guidance or even recommendations to stop the task to prevent injury. 

This scoring approach allows non-expert observers to indicate the degree of struggle/non-struggle. Note that we forced annotators to choose between struggle and non-struggle instead of allowing a neutral option. This avoided centring strategies on ambiguous judgements and allowed easy binarisation of annotations during early pilot approaches that classed only struggle presence/absence. Throughout our research, we found that indicating the degree of struggle is necessary to address significant challenges in classifying fine-grained features. 

As self-reporting levels of struggle can be unreliable and provide only a single judgement, we focus on a balance between expert and naive observers. We collected two distinct types of struggle determination annotations: naive `voter labels' obtained from Amazon Mechanical Turk (MTurk) participants and `expert annotations' (or `golden annotation' (GA)) obtained from the author who collected the video dataset, who repeatedly viewed them during analysis, and annotated action and error labels in the EPIC-Tent dataset. 
For each 10s video clip, there is only one expert annotation label, and there are multiple voters' labels that form a frequency distribution. This distribution consists of 15 or 20 voter labels. To extract a single struggle label from this distribution, we take the mode. Since voter labels and expert annotation come from distinct groups of individuals with differing prior knowledge, label distributions can vary across all video segments for each dataset. Consistency between the voters' modes and expert annotations indicates that the struggle-level annotations of the videos are more reliable.
In addition, the standard deviation of the voters' distributions provides a measure of the voter consistency or label ambiguity for a segment (smaller values indicate greater consistency, i.e. less label ambiguity). 
We explore these annotations in Section ~\ref{dataset statistics}. 

\subsubsection{Expert Annotator Qualifications and Training}
\label{expert-annotation-reliability}

Unlike the MTurk workers who only viewed 10-second clips for each task, the expert, as a paper author, had previously reviewed the full videos (multiple times) during data annotation and quality control. For expert annotation, this broader context helped ensure consistency, as the expert worked independently. In contrast, crowd annotators did not require prior task knowledge, but to achieve a relatively consistent struggle annotation, we relied on multiple crowd workers per clip and aggregated their inputs by averaging struggle levels or taking the mode after filtering out outliers. Additionally, the expert had prior experience coding actions, objects, and errors within the `tent' dataset, further honing their observation and annotation skills relevant to the 10-second clips.

We acknowledge the inherent ambiguity in struggle determination, similar to skill assessment. Much like Olympic judges offering slightly different scores for a performance of exceptional skill, we anticipate slight disagreement in struggle determination. In our dataset of short clips, struggle typically manifests as motor errors, such as abrupt stops, repetitive attempts without success, object drops, or awkward manipulation. We believe that basic motor struggle is readily detectable by most human observers based on the fluidity of movement. However, the expert's advantage lies in their extensive experience making such judgments, coupled with their unique knowledge of the full videos and specific task goals.


\begin{figure*}[ht]
    \centering
    \includegraphics[width=1.0\linewidth]{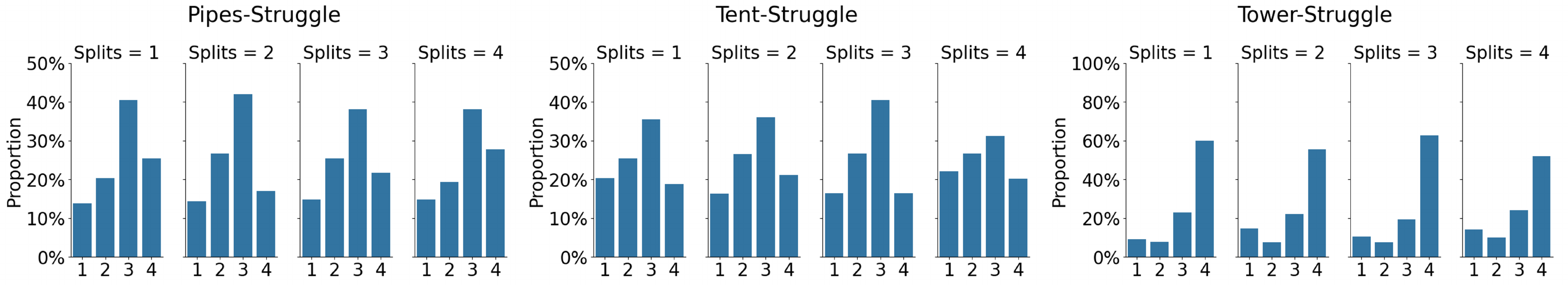}
    \caption{\textbf{Frequency distributions of four-level struggle labels across dataset splits.} The three graphs illustrate the distribution of struggle labels (ranging from 1 to 4 on the x-axis) for the activities Pipes-Struggle (left), Tent-Struggle (middle), and Tower-Struggle (right).}
    \label{fig:datasets-stats-splitlabels}
\end{figure*}

\begin{figure*}[ht]
    \centering
    \includegraphics[width=1.0\linewidth]{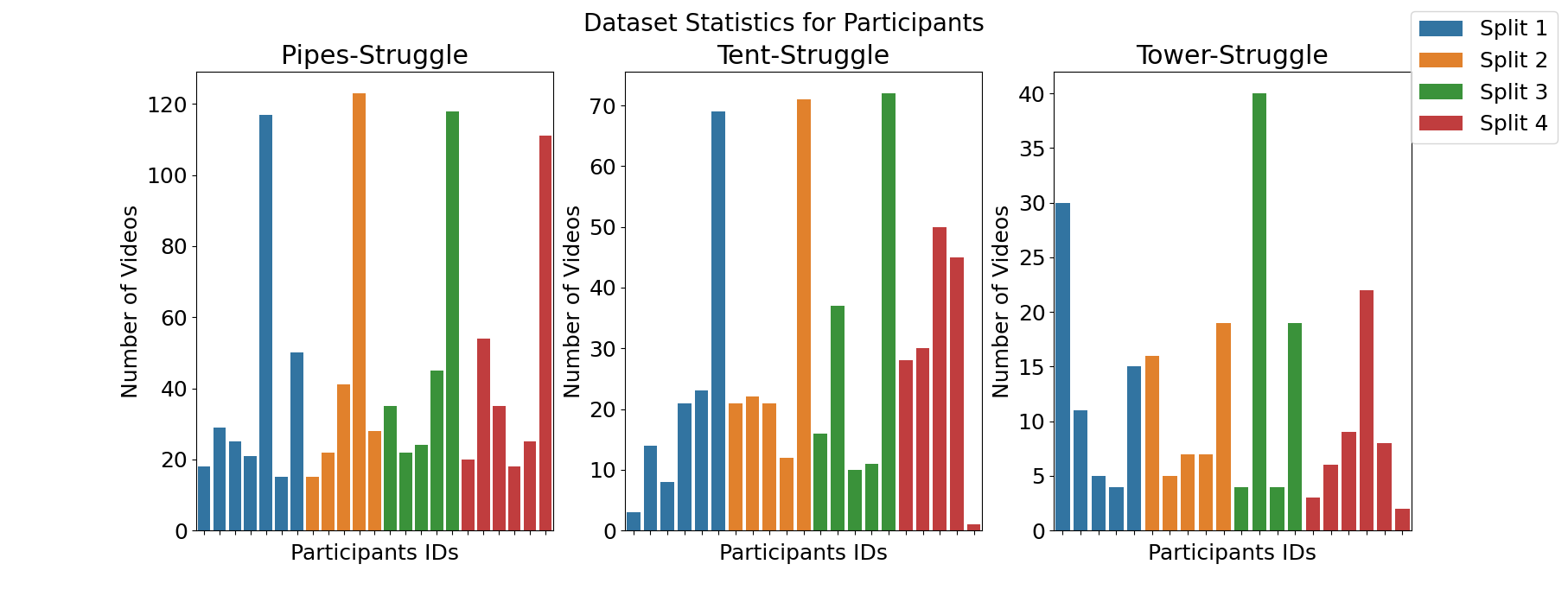}
    \caption{\textbf{A summary of the number of video samples corresponding to participants for each split.} The three graphs display participant IDs and the number of video segments for the activities Pipes-Struggle (left), Tent-Struggle (middle), and Tower-Struggle (right), respectively.}
    \label{fig:datasets-stats-participants}
\end{figure*}

\begin{figure*}[ht]
  \centering
  \includegraphics[width=1.0\linewidth]{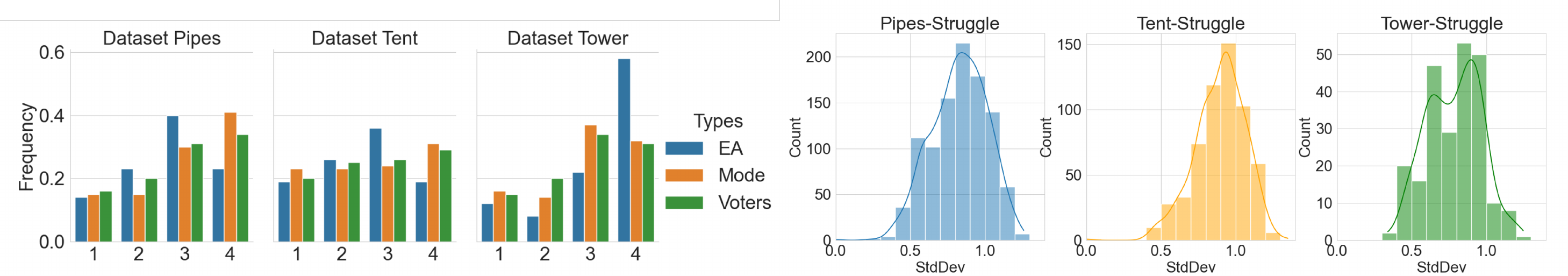}
  \caption{\textbf{Comparison of the distributions.} The first three graphs on the left display the frequency distribution of the struggle labels, including \textcolor{blue}{expert annotation (EA)}, \textcolor{orange}{voters mode (Mode)}, and \textcolor{green}{voters labels distributions (Voters)}, for the three activities. The three graphs on the right display the histogram with smoothed kernel density estimate (KDE) of the standard deviation (StdDev) of the voters' labels for the three activities.}
  \label{fig:dataset-stats-labels-stddev}
\end{figure*}

\begin{table*}[ht]
\scriptsize
\caption{\textbf{Statistics for the three activities}: the number of videos altogether and in each split, percentage of consistent videos in binary labels, percentage of consistent videos in four-class labels.}
\label{table:dataset statistics}
\centering
\begin{tabular}{cccc}
\toprule
Tasks & \#Videos (Split 1/2/3/4) & \%Binary Consistent & \%Four-way Consistent \\ 
\midrule
Pipes & 1011 (275/229/244/263)   & 80.61\%           & 45.40\%             \\ 
Tent  & 585 (138/147/146/154)    & 72.48\%           & 42.91\%             \\ 
Tower & 236 (65/54/67/50)        & 84.32\%           & 52.12\%             \\ 
\bottomrule
\end{tabular}
\end{table*}

\subsubsection{MTurk Annotator Validation}
\label{mturk-annotation-val}

We recognize that struggle determination and its annotation is an ambiguous task, which is part of the challenge and research value for this problem. However, we still want to ensure annotation by crowd-sourcing is as valid as possible and that the votes and MTurkers can be trusted. We use the following assumptions in this ambiguous undertaking: 1) we assume the expert annotator is trustworthy, 2) ambiguity between sides of struggle (e.g. definitely struggling and slightly struggling and similarly for non-struggling) is acceptable and expected, 3) If an MTurker is untrustworthy in one of the assignments, all results by that MTurker is questionable and best to remove. Based on these assumptions, we implemented the following validation checks:

\begin{itemize}
    \item If all votes for all videos assigned to the MTurker on an assignment are the same, all results from all assignments by that MTurker are rejected.
    \item We also identified 6 validation videos using three {\it definitely struggling} and three {\it definitely non-struggling} videos as labelled by the expert annotator. These videos were further checked for unanimous agreement by simple inspection by the authors.
    \item These 6 validation videos were included as part of the MTurk assignments, and for every MTurker, we calculate the level of disagreement w.r.t. the validation videos and the expert's votes. If the MTurker's votes are {\it on the same side} of struggle w.r.t the expert's vote, i.e. (struggling (3,4) or non-struggling (1,2)), this does not count as a disagreement. The MTurker is rejected if the ratio of disagreement w.r.t. the validation videos is $> 0.33$, i.e. three or more out of 6 disagreements for the validation videos.
\end{itemize}

\subsection{Struggle Statistics}
\label{dataset statistics}

The three subsets of Our struggle determination dataset, Tent-Struggle, Pipes-Struggle, and Tower-Struggle, contain 585, 1011, and 236 annotated 10s videos, respectively, totalling ~5.09 hours recorded with ~725,100 frames. We further divide each of the subsets into four splits for four-fold cross-validation such that the video samples from the same participants are within the same split and the struggle label distributions are similar across the four splits (see Figure~\ref{fig:datasets-stats-splitlabels} for more details). The final split of participants and the corresponding number of video segments they contributed are shown in Figure~\ref{fig:datasets-stats-participants}. We note that some participants have contributed a significantly higher number of video segments than the other participants. 

When considering struggle labels for video samples, there are three types of labels to take into account: a) Expert annotation---the struggling label given by the expert annotator; b) Voters' mode---the statistical mode's struggle value the voters give; and c) Voters' label distribution---composed of the frequencies of voters' labels on each of the struggle levels for each video sample. 

Table \ref{table:dataset statistics} shows the statistics for the three subsets, including the total number of videos, the number of videos per split and the proportion of videos whose voters' mode label is consistent with expert annotations. 

Figure \ref{fig:dataset-stats-labels-stddev} shows both the overall frequencies for each struggle label and the histograms of standard deviations that indicate the label ambiguity for each dataset. The histograms of standard deviations tell us that the Tent-Struggle dataset has the most ambiguous voter labels with a standard deviation between 0.9 and 1. This is followed by the Pipes-Struggle dataset and then the Tower-Struggle dataset, which has the least ambiguous labels. These ambiguities reflect the degree of variation innate to each task, with a well-defined task, such as Towers of Hanoi, having less ambiguity and assembling the non-rigid tent outdoors having the highest. The variation of ambiguity is a further element of interest for struggle-aware systems. Furthermore,  Figure \ref{fig-appendix:struggle_intragroup_disagreement} in the Appendix shows the intra-group voter disagreement in the dataset. This figure illustrates additional analysis on the frequency of annotators' disagreements regarding the binary struggle/non-struggle labels, focusing on the minority group, which has fewer instances of disagreement, ranging from 0 to 0.5. As shown in the figure, video samples with a minority group disagreement frequency of 0.5 constitute only a small proportion of the entire dataset.

\section{Experiments}
\label{sec: experiments-new}
In this section, we provide baseline results for our struggle determination dataset and structure experiments to answer the following questions: (i) What are the most important modelling approaches for struggle determination? (ii) To what extent can a struggle determination model being trained on one subset generalize to other struggle subsets? (iii) How do different backbone deep architectures impact struggle determination performance? and (iv) What are the most appropriate baseline models for struggle determination? 

\subsection{Comparison of Struggle Modelling Approaches}
\label{model-approaches}
\begin{table*}[ht]
\scriptsize
\caption{Comparison of Classification and Regression-to-Classification Accuracy based on MViTv2~\citep{li2022mvitv2}.}
\label{tab:cls-reg-comparison}
\centering
\begin{tabular}{lcccc}
\toprule
\multirow{2}{*}{\textbf{Activity}} & \multicolumn{2}{c}{\textbf{Classification}} & \multicolumn{2}{c}{\textbf{Regression-to-Classification}} \\
\cmidrule(lr){2-3} \cmidrule(lr){4-5}
                  & Binary Accuracy & Four-Way Accuracy & Binary Accuracy & Four-Way Accuracy \\
\midrule
Pipes-Struggle    & \textbf{76.68}         & \textbf{48.43}            & 75.03         & 44.71 \\
Tent-Struggle     & \textbf{64.42}         & \textbf{37.06}            & 59.86         & 35.60 \\
Tower-Struggle    & \textbf{84.29}         & \textbf{57.42}            & 81.29         & 33.41 \\
\bottomrule
\end{tabular}%
\end{table*}

We initially thought about three different modelling approaches to assess struggle determination comprehensively.
First, \textbf{struggle classification} includes binary and four-way classifications using expert labels. In the four-way classification, we use the original struggle labels ranging from 1 to 4, corresponding to definitely non-struggle to definitely struggle. In binary classification, we binarize the label of struggling versus non-struggling by grouping 1\&2 votes and 3\&4 votes. Secondly, \textbf{regression} fits the degree of struggle. Here we treat the struggle level labels from the experts as continuous real numbers ranging from 1 to 4. The third approach is \textbf{label distribution learning}. Given the video samples, we train a deep model that can predict the frequency distribution over the four struggle levels while using the frequency distributions of the voters' annotations for each video segment as the target. This modelling approach is a supplementary way of making full use of the voters' annotations, and thus, we will include it in our ablation study (Section~\ref{sec: label distribution learning}). 

Compared to struggle classification, struggle regression provides the advantage of modelling struggle levels as continuous real numbers from 1 to 4, capturing subtle variations more effectively. To determine the most suitable modelling approach, we first train a deep model for struggle classification and then train the same model for struggle regression. The regression outputs are then quantized into either binary or four-way struggle classifications, and we evaluate the accuracy rate for comparison. We adopt this regression-to-classification approach because the primary goal in struggle determination is to conclude whether a person is struggling or not, rather than merely predicting a numerical score.

The experimental results are presented in Table~\ref{tab:cls-reg-comparison}. We selected MViTv2~\citep{li2022mvitv2} for this comparison due to its efficiency in capturing spatiotemporal features as a recent representative transformer-based model. The results show that both binary and four-way struggle classification achieve higher accuracy when the model is trained as a classification task rather than a regression task. This trend holds consistently across all three struggle datasets, suggesting a significant advantage of training the deep model as a classification task. 
One possible reason is that the regression model predicts a continuous struggle level, and when these scores are quantized into binary or four-way classes, some instances may be misclassified due to threshold selection. 
We also conducted extensive experiments comparing classification and regression results across different deep models to assess the consistency of this trend (see Appendix Table~\ref{tab:tsn_comparison} and \ref{tab:slowfast_comparison}). The findings consistently indicate that models trained for classification outperform those trained for regression. 


\subsection{Generalisation Experiments}
\label{generalisation}
\begin{table*}[ht]
\caption{\textbf{Results show a comparison between training the model on a combined training set versus training on each of the struggle subsets separately.} `Best Epoch Over Three' refers to selecting a single epoch that performs relatively best based on the average accuracy across all three validation sets from all three struggle subsets, which is then evaluated individually on each subset. `Best Epochs Separately' refers to choosing the best epochs with the highest accuracy on each subset’s validation set. The results are reported using the mean accuracy across all cross-validation splits.
}
\label{tab:mvitv2-generalization-combinedsets}
\centering
\resizebox{\textwidth}{!}{%
\begin{tabular}{lcccccc}
\toprule
Dataset & \multicolumn{3}{c}{Test Sets - Binary Cls. Accuracy (\%)} & \multicolumn{3}{c}{Test Sets - Fourway Cls. Accuracy (\%)} \\ \cmidrule(lr){2-4} \cmidrule(lr){5-7}
Training Sets & Pipes-Struggle & Tent-Struggle & Tower-Struggle & Pipes-Struggle & Tent-Struggle & Tower-Struggle \\ \midrule
Pipes-Struggle & 76.68 & 49.22 & 71.56 & 48.43 & 21.58 & 40.79 \\
Tent-Struggle & 53.99 & 64.42 & 75.54 & 30.97 & 37.06 & 29.04 \\
Tower-Struggle & 62.31 & 47.90 & 84.29 & 24.31 & 25.42 & 57.42 \\ \midrule
Combined (Best epoch over three) & 77.29 & 66.71 & 90.22 & 48.79 & 40.26 & 66.55 \\
Combined (Best epochs Separately) & \textbf{79.83} & \textbf{69.48} & \textbf{91.74} & \textbf{51.89} & \textbf{41.81} & \textbf{68.81} \\
\bottomrule
\end{tabular}%
}
\end{table*}

After identifying struggle classification as a priority modelling approach for struggle determination, we conduct experiments to show the generalisation capability among the three different scenarios: Pipes-Struggle, Tent-Struggle, and Tower-Struggle. 
We used the same MViTv2~\citep{li2022mvitv2} model to conduct the generalisation experiment. 
We first trained and evaluated the model on the same struggle subset and then conducted zero-shot evaluations on the other two struggle subsets. See Table~\ref{tab:mvitv2-generalization-combinedsets} for the results. 
Although the struggle classification accuracy dropped when directly evaluated on unseen struggle subsets, some training-testing combinations demonstrate potential generalization. 
In general, models trained on Pipes-Struggle and Tent-Struggle tend to transfer better to Tower-Struggle, achieving relatively high accuracy. This is followed by models trained on Tent-Struggle and Tower-Struggle transferring to Pipes-Struggle, though the generalization is not as strong. 
The struggle in the Tent-Struggle subset appears to be the hardest to generalize to, as models trained on Pipes-Struggle and Tower-Struggle achieve the lowest accuracy when tested on Tent-Struggle. This may be due to the struggle patterns in Tent-Struggle being more complex and occurring in an outdoor setting, compared to the other two indoor environments. 
Additionally, models trained on Tower-Struggle struggle to generalize to the other subsets, especially in four-way classification, where accuracy drops significantly. This may be because the struggle patterns in the Tower-Struggle subset are relatively simple, the task complexity is low, and the number of data samples is the smallest among the three subsets. 
These findings suggest that certain struggle features are more transferable across subsets, while others remain more dataset-specific.

We also conducted multi-activity learning experiments to train the model using a combination of the training splits of the three struggle subsets and evaluating the model on the validation splits individually. We concatenate the training data to train the MViTv2~\citep{li2022mvitv2} model and run three evaluation stages separately to evaluate the accuracy rate on the three subsets. 
We present results in Table~\ref{tab:mvitv2-generalization-combinedsets}, where we report both the binary classification accuracy and the four-way classification accuracy by comparing the combined training results with the baseline results of training separately. The results are reported in two ways: (1) selecting the epochs with the highest accuracy on each validation set individually, which may be different for each set, and (2) selecting a single epoch that provides the best average accuracy across all three validation sets. 
As shown in Table~\ref{tab:mvitv2-generalization-combinedsets}, training the MViTv2~\citep{li2022mvitv2} model on combined struggle subsets demonstrates a significant trend of improvement in both binary and four-way classification accuracy, with increases generally ranging from 3\% to 11\% compared with training on each of the struggle subsets separately. This is not only because the MViTv2~\citep{li2022mvitv2} model benefits from having more training data for struggle classification, but also because it may indicate that struggle patterns are potentially transferable across different task-performing activities.
We show that the generalisation across datasets is challenging for MViTv2~\citep{li2022mvitv2}, particularly in a zero-shot setting. Our results indicate that the struggle classification accuracy drops significantly when the model is tested on a different struggle subset. Moreover, MViTv2~\citep{li2022mvitv2} requires a large amount of data to generalise well or, more specifically, to surpass the accuracy achieved when trained and tested on individual struggle subsets. This is evident from the bottom two rows of our results, where training on combined subsets leads to better performance on the evaluation test sets.

\subsection{Model Comparison on Struggle Classification}
\label{model-comparison}
\subsubsection{Baseline Methods}
\label{sec: baseline methods} 

We compare four different types of deep architectures in our struggle classification experiments: (1) 2D-ConvNet, represented by Temporal Segment Networks (TSN) \citep{wang2016temporal}, is implemented using \cite{tsn-pytorch}. (2) 3D-ConvNet, represented by SlowFast Networks \citep{feichtenhofer2019slowfast}, is based on ResNet50 (SlowFast-R50) \citep{fan2020pyslowfast}. (3) Hybrid architectures, which combine 3D-ConvNets and Vision Transformers \citep{https://doi.org/10.48550/arxiv.2010.11929}, include VTN-SlowFast-ViT and SlowFast-gMLP. (4) A Transformer-based model, MViTv2 \citep{li2022mvitv2}, which has been adopted in the previous experiments.
\added{Besides, we included the Video-LLaVA~\citep{lin2023video} VLM model, evaluated in a zero-shot setting without further fine-tuning, as shown in Appendix~\ref{add-vlm}.}

The two-hybrid architectures, VTN-SlowFast-ViT and SlowFast-gMLP, utilize the SlowFast Networks as the backbone to extract spatial-temporal features from video sequences. These features are then processed by different attention mechanisms: VTN-SlowFast-ViT incorporates a Vision Transformer encoder block \citep{https://doi.org/10.48550/arxiv.2010.11929}, implemented using \cite{vit-pytorch}, while SlowFast-gMLP applies a gMLP layer \citep{DBLP:journals/corr/abs-2105-08050}, implemented using \cite{gmlp-pytorch}.

Specifically, given a dataset containing a set of $K$ videos $P=\{p_i, 1 \leq p_i\leq K\}$. We sample $N$ frames from each video sample $p_i$ and feed them into the SlowFast backbone $\Phi (\cdot)$. This process result in $N/8$ feature vectors, which are then passed through either the Vision Transformer encoder or a gMLP layer. Finally, an MLP head is applied for struggle classification. 

We employ four-fold cross-validation to train and evaluate these deep models across all tasks and report the averaged top-1 classification accuracy over all four validation splits. To mitigate overfitting, we freeze the parameters in the backbone layers during training, effectively performing linear probing for TSN~\citep{wang2016temporal} and SlowFast-R50~\citep{feichtenhofer2019slowfast}. For the hybrid architectures, only the Vision Transformer-based layers are trained while keeping the backbone frozen.

\begin{table*}[ht]
\caption{\textbf{Struggle classification results.} Classification accuracy for binary classification, four-way classification and the four-way to binary classification. 
}
\label{table:classification results}
\centering
\resizebox{\textwidth}{!}{%
\begin{tabular}{@{}lccccccccccc@{}}
\toprule
\multirow{2}{*}{Dataset}                                   & \multicolumn{3}{c}{Pipes-Struggle}                                                               & \multicolumn{1}{c}{} & \multicolumn{3}{c}{Tent-Struggle}                                                           & \multicolumn{1}{c}{} & \multicolumn{3}{c}{Tower-Struggle}                                                          \\ 
                                           & \multicolumn{3}{c}{Top-1 Accuracy (\%)}                                                          & \multicolumn{1}{c}{} & \multicolumn{3}{c}{Top-1 Accuracy (\%)}                                                     & \multicolumn{1}{c}{} & \multicolumn{3}{c}{Top-1 Accuracy (\%)}                                                     \\ \cmidrule(lr){2-4} \cmidrule(lr){6-8} \cmidrule(l){10-12} 
Models                                     & \multicolumn{1}{c}{Binary} & \multicolumn{1}{c}{Four-Way} & \multicolumn{1}{c}{Four-to-Bin} & \multicolumn{1}{c}{} & \multicolumn{1}{c}{Binary} & \multicolumn{1}{c}{Four-Way} & \multicolumn{1}{c}{Four-to-Bin} & \multicolumn{1}{c}{} & \multicolumn{1}{c}{Binary} & \multicolumn{1}{c}{Four-Way} & \multicolumn{1}{c}{Four-to-Bin} \\ \midrule
Random                                     & 53.38                        & 28.62                           & 53.38                           &                      & 50.59                      & 27.12                        & 50.59                           &                      & 67.86                      & 40.45                        & 67.86                           \\
Majority Class                             & 62.60                        & 39.60                           & 62.60                           &                      & 54.91                      & 35.79                        & 54.91                           &                      & 79.74                      & 57.56                        & 79.74                           \\
Voters' Baseline                             & 80.56                        & 45.31                           & 80.56                           &                      & 72.63                      & 43.12                        & 72.63                           &                      & 84.41                      & 51.86                        & 84.41                           \\ \midrule
TSN \citep{wang2016temporal}          & 74.54                        & 48.59                           & 71.43                           &                      & 68.05                      & 38.23                        & 61.24                           &                      & 82.59                      & 59.49                        & 78.68                           \\
SlowFast-R50 \citep{feichtenhofer2019slowfast} & 78.01                        & \textbf{50.77}                  & \textbf{75.73}                  &                      & \textbf{69.56}             & 39.48                        & \textbf{65.19}                  &                      & \textbf{88.24}             & 63.92                        & 80.27                           \\
SlowFast-gMLP                              & \textbf{78.59}               & 49.70                           & 74.49                           &                      & 68.00                      & 39.52                        & 64.68                           &                      & 87.86                      & \textbf{66.40}               & \textbf{83.17}                  \\
VTN-SlowFast-ViT                           & 77.80                        & \textbf{50.83}                  & 72.48                           &                      & 68.53                      & \textbf{40.11}               & 60.86                           &                      & \textbf{88.11}             & 64.03                        & 82.54                           \\ 
MViTv2 \citep{li2022mvitv2} & 76.68                        & 48.43                  & 72.25                  &                      & 64.42             & 37.06                        & 58.15                 &                      & 84.29             & 57.42                        & 82.59                           \\ \bottomrule
\end{tabular}%
}
\end{table*}

\subsubsection{Results and Model Comparison}
The struggle classification results are shown in Table~\ref{table:classification results}. 
The first three rows represent naive baselines: (1) the random performance baseline, which uses the frequency of classes as the probability for random predictions; (2) the majority class baseline, which predicts the majority class and reports its percentage; and (3) the voter's baseline, where the mode of the voters' annotations is used to compare with the expert annotations and calculate the averaged accuracy. 
The experimental results indicate a performance gap between the voter’s baseline accuracy and the best-performing deep models.

Among the evaluated models, TSN~\citep{wang2016temporal} and MViTv2~\citep{li2022mvitv2} exhibit relatively low accuracy in both binary and four-way classifications. The poor performance of the TSN~\citep{wang2016temporal} in struggle classification is probably attributed to the 2D-ConvNet design, which lacks explicit temporal modelling and relies solely on global average pooling (GAP) to aggregate frame-level features. This approach disregards frame order, negatively impacting performance. 
On the other hand, MViTv2~\citep{li2022mvitv2}, despite leveraging spatial-temporal dependencies through multi-head self-attention, still underperforms. This may be due to the fact that transformer-based models typically require large amounts of training data to effectively capture useful patterns. This hypothesis is supported by the significant improvement in struggle classification performance when MViTv2~\citep{li2022mvitv2} is trained on combined datasets, as shown in Table~\ref{tab:mvitv2-generalization-combinedsets}, which also outperforms the struggle classification accuracy of all other deep models shown in Table~\ref{table:classification results}.

In contrast, 3D-ConvNets, represented by the SlowFast~\citep{feichtenhofer2019slowfast} and its enhanced variants like VTN-SlowFast-ViT and SlowFast-gMLP, which incorporate more advanced temporal modelling, generally achieve higher accuracy. This suggests that the SlowFast model, as a feature extraction backbone, is more robust for struggle classification, particularly when training data is limited. The struggle classification accuracy is similar across the three models that use SlowFast~\citep{feichtenhofer2019slowfast} as the backbone. This suggests that incorporating MLP- or Transformer-based architectures for additional temporal modelling does not significantly improve performance. A likely explanation is that our video-level weak labels for video segments do not require extensive temporal processing beyond the initial spatial-temporal feature extraction. As a result, the potential of the MLP/Transformer layers to enhance accuracy remains limited in this context.

\subsection{Implementation Details}
We conducted two classification experiments: binary classification, which distinguishes between struggle and non-struggle, and four-class classification, which predicts the level of struggle. In both cases, the model is supervised using the Cross-Entropy (CE) loss. Top-1 accuracy is used as the evaluation metric. During training, we uniformly sample 128 frames from each video and randomly crop the input frames from $256\times456$ to $256\times256$. Additionally, a horizontal flip is applied with a probability of 0.5 during training. For the evaluation stage, we perform 5 random crops and ensemble the predictions to improve robustness.

For TSN~\citep{wang2016temporal}, we implement the BN-Inception base architecture with the average consensus for the output from each segment. We use ImageNet \citep{ILSVRC15} pre-trained features and train only the fully connected layer for classification. To address overfitting, we set the dropout rate to 0.5. The model is trained for 65 epochs using mini-batch stochastic gradient descent with a momentum of 0.9 and a weight decay of $5\times10^{-4}$. The learning rate starts at 0.0001 and decreases by a factor of $10\times$ at epochs 30 and 60.

For the SlowFast~\citep{feichtenhofer2019slowfast} model and its variants, we use the architecture that is based upon 3D-ResNet50 and load parameters pre-trained on Kinetics-400~\citep{https://doi.org/10.48550/arxiv.1705.06950}. During training, we freeze the backbone parameters and train the remaining MLP layers. The same optimizer is used, but the weight decay is set to $1\times10^{-5}$. We train these models for 70 epochs with a cyclic cosine decay learning rate scheduler~\citep{athiwaratkun2019consistent}, where the learning rate starts at 0.001, decreases by a factor of 100 over 50 epochs, and includes two restart intervals of 10 epochs each. During these intervals, the learning rate increases by a factor of 10 at the start and decreases back to the minimum value.

For MViTv2~\citep{li2022mvitv2}, we implement its small variant to accommodate limited GPU memory. We load parameters pre-trained on Kinetics-400~\citep{https://doi.org/10.48550/arxiv.1705.06950} and fine-tune the model using linear probing to mitigate overfitting. The AdamW optimizer is used with a weight decay of $5\times10^{-4}$ and a cyclic cosine decay learning rate scheduler. The first 10 epochs serve as a warm-up stage, starting at 0.1 times the base learning rate of 0.001. The learning rate is then gradually decreased by a factor of 0.01 from epoch 10 to epoch 50, followed by two fixed restart intervals where the learning rate decreases from 0.1 times the base learning rate to 0.01 times the base learning rate. The total training duration is 70 epochs. In the generalization experiments, we maintain the same settings but employ a multi-step learning rate scheduler. The base learning rate starts at $1\times10^{-6}$ and increases by a factor of 10 at epochs 30 and 60, with the models trained for 70 epochs for full parameter fine-tuning.

All experiments are conducted using two Titan X (Pascal) 12GB GPUs or two GTX 1080Ti 11GB GPUs.

\section{Ablation Study}
\label{sec: ablation}

\begin{figure*}[ht]
  \centering
  \includegraphics[width=1.\linewidth]{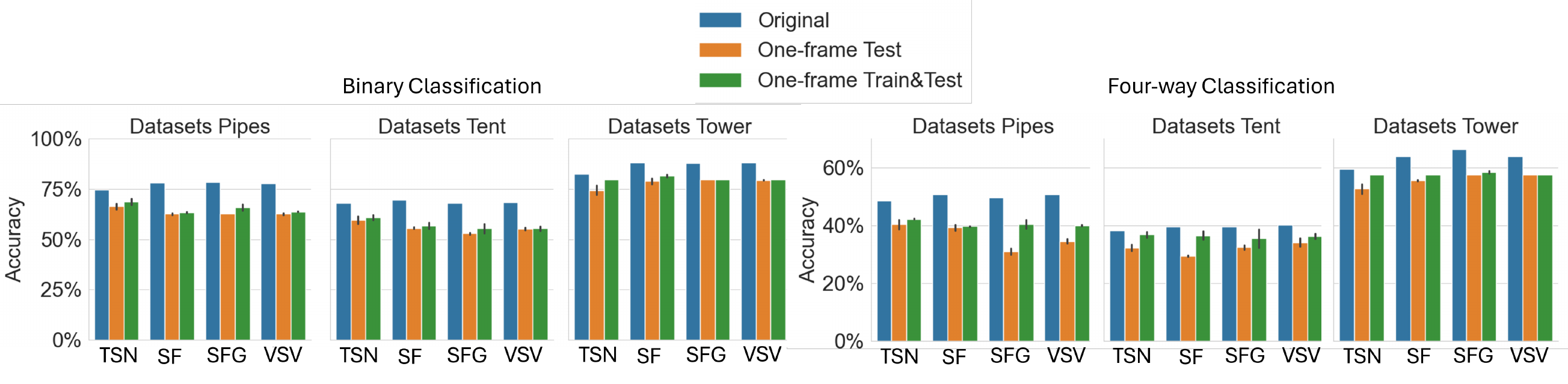}
  \caption{\textbf{Importance of Multiple Frames.} The left three bar charts show binary classification results for Pipes-Struggle, Tent-Struggle, and Tower-Struggle, while the right three show four-way classification for the same activities. Each group of bars represents the base model (blue), evaluation with a single frame (orange), and training/evaluation with a single frame (green). Error bars indicate the standard deviation (StdDev) across sampled frames from 25\%, 50\%, and 75\% of the video. See Appendix Table~\ref{table:one frame} for detailed result numbers. Model abbreviations: TSN~\citep{wang2016temporal}, SF: SlowFast~\citep{feichtenhofer2019slowfast}, SFG: SlowFast-gMLP, VSV: VTN-SlowFast-ViT.}
  \label{fig:one-frame}
\end{figure*}

\begin{figure*}[ht]
  \centering
  \includegraphics[width=1.0\linewidth]{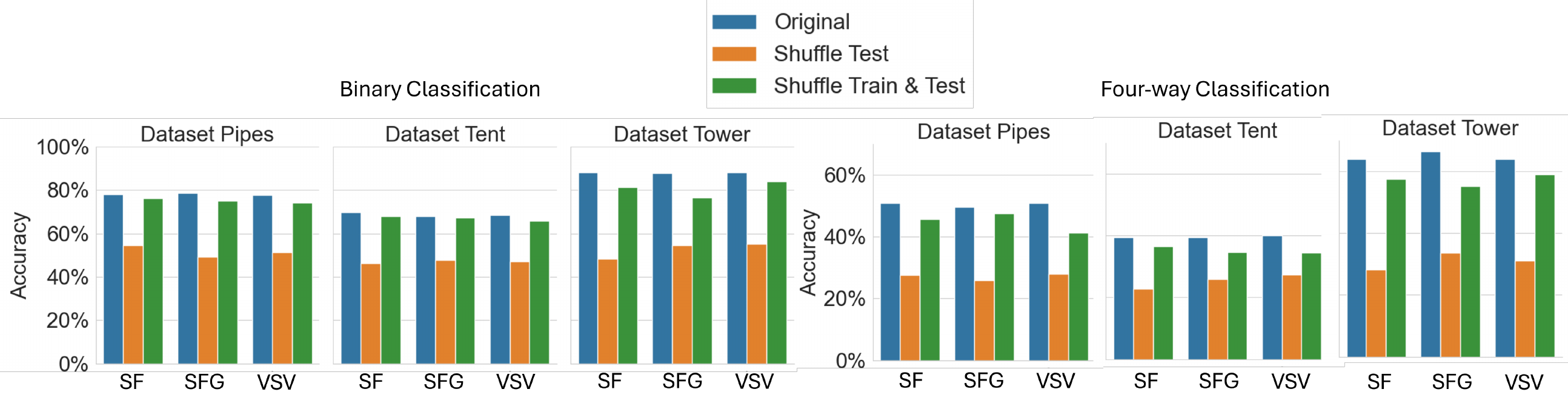}
  \caption{\textbf{Importance of Temporal Ordering.} The left three bar charts show binary classification results for Pipes-Struggle, Tent-Struggle, and Tower-Struggle, while the right three show four-way classification for the same activities. Each group of bars represents input frames in the correct order (blue), evaluation with shuffled frames (orange), and training/evaluation with shuffled frames (green). See Appendix Table \ref{table:shuffle frame} for detailed result numbers. Model abbreviations: SF: SlowFast~\citep{feichtenhofer2019slowfast}, SFG: SlowFast-gMLP, VSV: VTN-SlowFast-ViT.}
  \label{fig:shuffle-frames}
\end{figure*}

In this section, we investigate the importance of temporal modelling, motion understanding, and alternative training strategies in struggle determination.
Specifically, we conduct two series of experiments: (1) validating the use of multiple frames for struggle determination by comparing it with single-frame input and (2) evaluating the importance of temporal ordering in frame sequences for struggle determination.
Additionally, we explore alternative training strategies by presenting baseline results for struggle regression and struggle label distribution learning. We also compare different deep learning models to identify the most effective architecture for struggle determination under these two training approaches.

\subsection{Multiple Frames}
To demonstrate the importance of using multiple frames, we conducted experiments comparing a single-frame input to the original multi-frame setting.

First, we initialized the model with parameters trained on 128 frames, which is the standard setting for our experiments. We then tested the model using only one frame from each video sequence. The selected frame was taken from three different positions within each video segment—at 25\%, 50\%, and 75\% of its duration. Since 3D-ConvNets require a minimum number of input frames, we duplicated the selected frame to meet these requirements: 8 frames for SlowFast~\citep{feichtenhofer2019slowfast} and 128 frames for SlowFast-gMLP and VTN-SlowFast-ViT.

To ensure consistency between training and testing, we conducted an additional experiment where we trained the model using randomly selected single frames from each video. During testing, we applied the same single-frame selection strategy as before, choosing frames from 25\%, 50\%, and 75\% of each video segment.

The results of these two approaches are shown in Figure~\ref{fig:one-frame}. We report the average test accuracy for single-frame inputs across the three proportions. Across all tested models and activities, we observed a significant drop in accuracy when using only one frame. Since our dataset consists of randomly trimmed 10-second video segments, accuracy values for frames sampled from 25\%, 50\%, and 75\% are expected to be similar, leading to a small standard deviation (StdDev) in accuracy, as reflected in the bar plots.

Even when the models were trained using a single frame, the test accuracy remained notably lower compared to the multi-frame setting, though slightly higher than testing on a single frame without training adjustments. This suggests that a single frame does not provide sufficient visual information for deep learning models to reliably determine whether a person is struggling. 

Therefore, effectively capturing the useful temporal motion features is crucial for our struggle determination task. A sequence of frames can typically reveal anomalous hand motions such as `unsmooth' motions, stopping, and repeating that indicate the person did not have things under control. A single frame cannot represent these dynamic features. 

\subsection{Temporal Ordering}

\begin{table*}[ht]
\caption{\textbf{Struggle regression results}. Mean Squared Error (MSE) and Mean Absolute Error (MAE), along with converted binary and four-way classification accuracy.
}
\label{table:regression results}
\centering
\resizebox{\textwidth}{!} 
{
\begin{tabular}{clcccc}
\toprule
\multirow{2}{*}{Activities} & \multirow{2}{*}{Models} & \multicolumn{4}{c}{Regression \&   Classification} \\ \cmidrule{3-6}
                            &                  & MSE$\downarrow$   & MAE$\downarrow$   & Binary   Acc.$\uparrow$& Four-way   Acc.$\uparrow$\\
\midrule
\multirow{4}{*}{Pipes-Struggle} & Voters' Baseline & 0.890 & 0.654 & 80.56\% & 45.31\% \\ \cmidrule{2-6}
                            & TSN \citep{wang2016temporal}              & 0.798 & 0.725 & 72.42\%       & 45.02\%         \\
                            & SlowFast-R50 \citep{feichtenhofer2019slowfast}    & \textbf{0.645} & 0.655 & 76.89\%       & 45.28\%         \\
                            & VTN-SlowFast-ViT & 0.648 & \textbf{0.650}  & \textbf{77.30\%}       & \textbf{48.56\%}         \\
                            & MViTv2 \citep{li2022mvitv2} & 0.745 & 0.708  & 75.03\%       & 44.71\%         \\
\midrule
\multirow{4}{*}{Tent-Struggle} & Voters' Baseline & 1.031 & 0.713 & 72.63\% & 43.12\% \\ \cmidrule{2-6} 
                            & TSN \citep{wang2016temporal}             & 0.888 & 0.768 & 68.27\%       & 40.95\%         \\
                            & SlowFast-R50 \citep{feichtenhofer2019slowfast}    & \textbf{0.850}  & \textbf{0.760}  & \textbf{69.06\%}       & \textbf{41.63\%}         \\
                            & VTN-SlowFast-ViT & 0.868 & 0.770  & 66.39\%       & 40.69\%         \\
                            & MViTv2 \citep{li2022mvitv2} & 1.055 & 0.858  & 59.86\%       & 35.60\%         \\
\midrule
\multirow{4}{*}{Tower-Struggle} & Voters' Baseline & 0.726 & 0.561 & 84.41\% & 51.86\% \\ \cmidrule{2-6}
                            & TSN \citep{wang2016temporal}           & 0.935 & 0.748 & 85.35\%       & 58.01\%         \\
                            & SlowFast-R50 \citep{feichtenhofer2019slowfast}    & 0.793 & 0.690  & 86.89\%       & 56.00\%         \\
                            & VTN-SlowFast-ViT & \textbf{0.723} & \textbf{0.673} & \textbf{87.86\%}       & \textbf{61.73\%}         \\
                            & MViTv2 \citep{li2022mvitv2} & 0.955 & 0.828  & 81.29\%       & 33.41\%         \\
\bottomrule
\end{tabular}%
}
\end{table*}

To further examine the role of temporal motion features in determining struggle, we conducted a shuffle-frame test to assess the importance of frame ordering. 
In this experiment, we excluded TSN~\citep{wang2016temporal}, as it only considers 2D spatial features and averages the output logits of all input frames, making it inherently insensitive to frame order. 

We began by loading model parameters pre-trained on 128 frames in their correct sequential order. Then, for each video sample, we randomly shuffled the same number of 128 frames, disrupting their natural temporal sequence. The model was run in inference mode ten times, and we reported the average accuracy across these runs.

As shown in Figure~\ref{fig:shuffle-frames}, the accuracy dropped significantly to a near-random level, indicating that the models for determining struggle are highly dependent on the correct frame order. To ensure a fair comparison, we conducted an additional experiment in which we trained the models using shuffled frames and then evaluated them under the same conditions. While training on shuffled frames led to improved performance compared to testing on shuffled frames alone, the accuracy remained lower than when the models were trained and tested on correctly ordered frames.

These results suggest that deep learning models learn crucial temporal dependencies when trained on properly ordered sequences to determine struggle. When this temporal structure is disrupted, model performance deteriorates, confirming that the models rely on motion continuity to infer struggle-related cues. 

\subsection{Alternate Training Strategies}
This section explores alternative training strategies for struggle determination, focusing on struggle regression and struggle label distribution learning. \textbf{Struggle regression} provides a full regression baseline, complementing the regression-to-classification approach. \textbf{Struggle label distribution learning} trains models to predict the distribution of voters' annotations, making better use of crowd-sourced struggle labels. We detail the implementation of these strategies and evaluate their effectiveness across different deep models.

\subsubsection{Struggle Regression}
\label{sec: regression}

\paragraph{Implementation Details}
In the regression task, struggle labels are treated as continuous scores ranging from 1 (definitely non-struggle) to 4 (definitely struggle), representing the degree of struggle. A fully connected layer is added for regression, which takes the feature vectors as input and outputs a regression score. The predicted score is supervised using Mean Squared Error (MSE) with the target regression labels. Evaluation metrics include MSE and Mean Absolute Error (MAE) for regression, as well as top-1 accuracy for binary and four-way classification. To ensure a solid conclusion on whether a person is struggling and to what extent, the continuous regression scores are quantized into struggle-level categories by thresholds when computing classification accuracy for the struggle regression model. The hyperparameter settings for optimization remain the same as in the struggle classification task.

\paragraph{Regression Results} 
Regression results are shown in Table~\ref{table:regression results}. Overall, VTN-SlowFast-ViT achieves the lowest MSE and MAE and the highest binary and four-way classification accuracy for the Pipes-Struggle and Tower-Struggle activities. 
SlowFast-R50~\citep{feichtenhofer2019slowfast} follows closely, showing competitive regression performance and classification accuracy, particularly for Tent-Struggle. 
This suggests that incorporating a vision transformer layer on top of the frame-level features from the SlowFast~\citep{feichtenhofer2019slowfast} backbone as a temporal model improves the fit to the struggle regression labels.
In contrast, TSN~\citep{wang2016temporal} and MViTv2~\citep{li2022mvitv2} generally underperform compared to these two models.

\subsubsection{Struggle Label Distribution Learning}
\label{sec: label distribution learning}

\begin{table*}[ht]
\caption{\textbf{Struggle Label distribution learning results}. Mean Absolute Error (MAE) and Spearman's Rank Correlation (Spearman's Rho), along with the converted binary classification accuracy and four-way classification accuracy.
}
\label{table:ldl results}
\centering
\resizebox{\textwidth}{!} 
{
\begin{tabular}{clcccc}
\toprule
\multirow{2}{*}{Activities} & \multirow{2}{*}{Models} & \multicolumn{4}{c}{Label Distribution Learning} \\ \cmidrule{3-6}
                            &                  & MAE$\downarrow$  & Spearman's   Rho$\uparrow$ & Binary   Cls.$\uparrow$& Four-way   Cls.$\uparrow$\\
\midrule
\multirow{3}{*}{Pipes-Struggle} & TSN \citep{wang2016temporal}            & 0.13 & 0.5120            & 79.72\%       & 46.01\%         \\
                            & SlowFast-R50 \citep{feichtenhofer2019slowfast}    & 0.12 & 0.5614           & 79.84\%       & \textbf{48.79\%}         \\
                            & VTN-SlowFast-ViT &\textbf{0.12} &\textbf{0.5689}           & \textbf{80.30\%}       & 46.47\%         \\
                            & MViTv2 \citep{li2022mvitv2} &\textbf{0.12} & 0.5331           & 78.88\%       & 47.72\%         \\
\midrule
\multirow{3}{*}{Tent-Struggle}  & TSN \citep{wang2016temporal}            & 0.17 & 0.2307           & 54.04\%       & 28.41\%         \\
                            & SlowFast-R50 \citep{feichtenhofer2019slowfast}    & 0.14 & 0.2887           & 57.07\%       & 31.09\%         \\
                            & VTN-SlowFast-ViT & \textbf{0.13} & \textbf{0.3318}           & 63.29\%       & 34.95\%         \\
                            & MViTv2 \citep{li2022mvitv2} & \textbf{0.13} & 0.2730           & \textbf{64.58\%}       & \textbf{37.16\%}         \\
\midrule
\multirow{3}{*}{Tower-Struggle} & TSN \citep{wang2016temporal}             & 0.24 & 0.4168           & 71.06\%       & 37.06\%         \\
                            & SlowFast-R50 \citep{feichtenhofer2019slowfast}    & 0.15 & 0.4936           & 74.86\%       & 40.44\%         \\
                            & VTN-SlowFast-ViT &\textbf{0.13} &\textbf{0.5689}           & 77.58\%       & 50.20\%         \\
                            & MViTv2 \citep{li2022mvitv2} & 0.14 & 0.5264           & \textbf{ 80.27\%}       & \textbf{50.81\%}         \\
\bottomrule
\end{tabular}%
}
\end{table*}

\paragraph{Implementation Details}
In label distribution learning, we train a deep model to predict the frequency distribution of voters' struggle labels given video samples. In our implementation, the final layer output of the model has four nodes corresponding to each of the four struggle levels ranging from 1 (definitely non-struggle) to 4 (definitely struggle), followed by a SoftMax layer so that each node's output value ranges from 0 to 1 with the summation equalling to one. 

We optimize the Kullback–Leibler (KL) divergence loss function to align the predicted distribution with the frequency distribution of the voters' struggle labels for each video sample. For evaluation, we calculate Spearmans' Rank Correlation (Spearmans' Rho) between the predicted frequency distribution and the voters' struggle label frequency distribution so that the value ranges from -1 to 1, with a higher score indicating that the predicted distribution better aligns with its target. We also calculate the Mean Absolute Error (MAE) between the predicted and the target distribution. 
 
The converted classification accuracy is derived from the predicted voters' label frequency distribution. To convert the frequency distribution into class labels, we select the top-1 frequency label for four-way classification. For binary classification, we merge the predicted frequencies of labels 1 and 2 into the ``non-struggle" category and labels 3 and 4 into the ``struggle" category. We determine the accuracy by comparing the predicted highest-frequency label with the voters' mode label. 

The hyperparameter settings remain consistent with those used in previous tasks.

\paragraph{Distribution Results} 
The results for label distribution learning are shown in Table \ref{table:ldl results}. Due to the continuous nature of MAE and Spearman’s Rho evaluation metrics, discrepancies can be observed among the implemented baseline models. These discrepancies follow the same trend when converted to accuracy rates.

The MViTv2~\citep{li2022mvitv2} model achieves the highest binary and four-way classification accuracy rates on Tent-Struggle and Tower-Struggle. However, it does not perform as well in struggle label distribution regression, as indicated by its lower Spearman’s Rho and MAE scores. In contrast, VTN-SlowFast-ViT consistently achieves the best label distribution regression performance, with the lowest MAE and highest Spearman’s Rho across all activities. The converted classification accuracy is also among the highest although slightly lower than that of MViTv2~\citep{li2022mvitv2} on Tent-Struggle and Tower-Struggle. Additionally, SlowFast-R50~\citep{feichtenhofer2019slowfast} outperforms TSN~\citep{wang2016temporal} in both struggle label distribution regression and classification, showcasing that 3D-ConvNets is more advanced in extracting spatial-temporal features than the 2D-ConvNets. VTN-SlowFast-ViT further surpasses SlowFast-R50~\citep{feichtenhofer2019slowfast} in general, suggesting that its additional ViT layer helps better align predictions with the voters' label distributions.

\begin{figure*}[ht]
  \centering
  \includegraphics[width=0.9\linewidth]{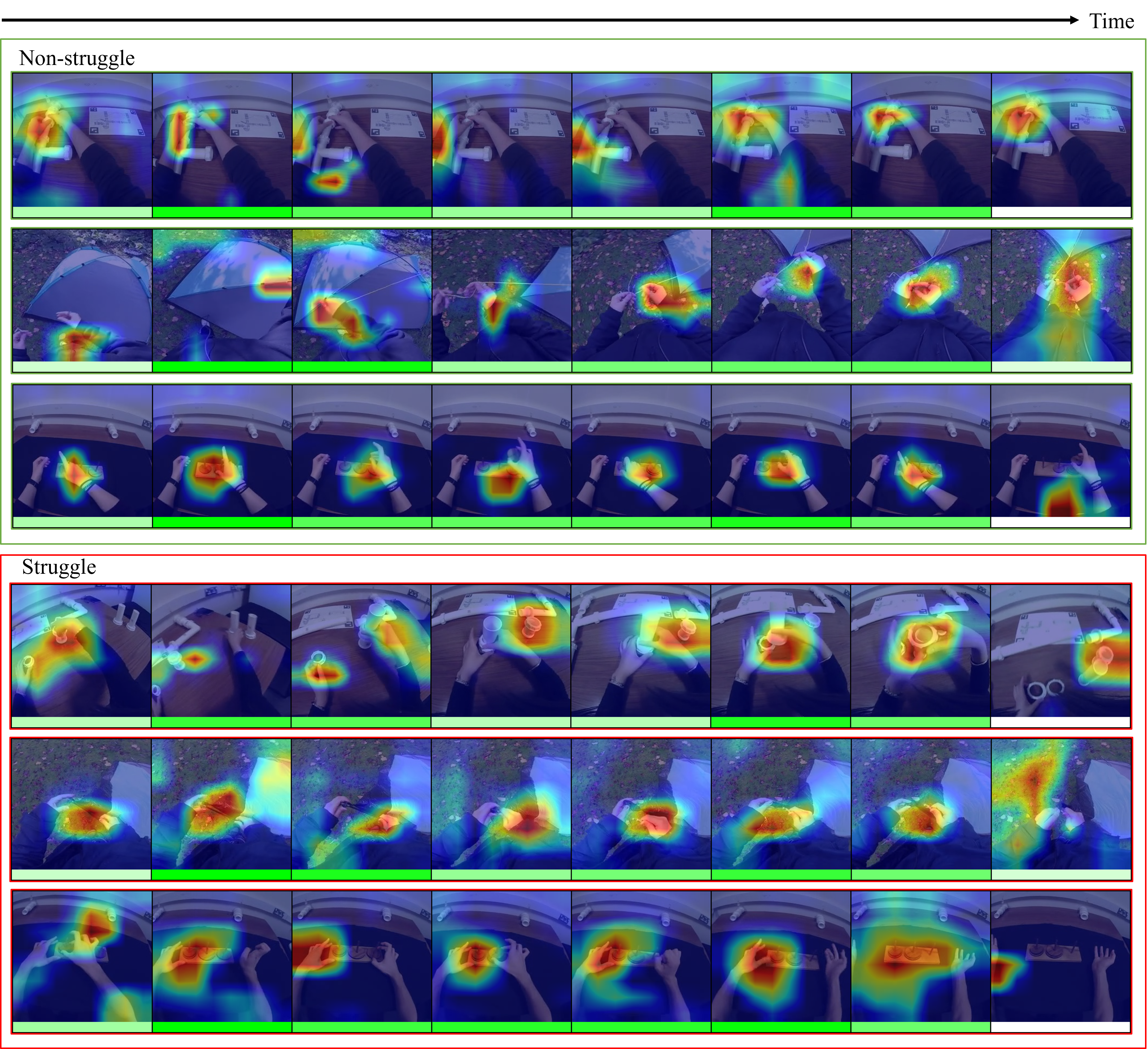}
  \caption{\textbf{Visualization results of the activation heatmaps and temporal attention scores}. Visualization results for both non-struggle (\textcolor{lime}{upper box}) and struggle (\textcolor{red}{bottom box}) are shown from activities: Pipes-Struggle (top), Tent-Struggle (middle), and Tower-Struggle (bottom) respectively with each row corresponding to the frames sampled from a video sample that is fed into the slow pathway and down-sampling into 8 frames. The heatmaps are the Grad-CAM++ \cite{DBLP:journals/corr/abs-1710-11063} visualization and the green bars below are the colourmap visualizations for the temporal attention scores of the corresponding video segments. The intensity of the green indicates the value of the attention score range from 0 to 1. The closer to 1, the darker the green colour.}
  \label{fig:vis}
\end{figure*}

\section{Attention Visualization}
\label{sec: visualization}
In this section, we visualize what the models focus on using attention weights and activation maps \citep{DBLP:journals/corr/abs-1710-11063}, ensuring they effectively detect participants' struggles. Specifically, we examine activation maps from the backbone network and attention scores from the self-attention layer of the VTN-SlowFast-ViT model.

Figure \ref{fig:vis} presents heatmaps from the final convolutional layer of the backbone network, generated using Grad-CAM++ \citep{DBLP:journals/corr/abs-1710-11063}, along with attention scores visualized as green bars. We provide examples from three activities: Pipes-Struggle, Tent-Struggle, and Tower-Struggle. The heatmaps reveal that the model predominantly focuses on hands and active objects, though occasionally not robust enough, causing attention to drift toward the background. For example, in the Pipes-Struggle visualised frames, the heatmaps highlight the hand of the person holding the pipes as well as the pipes that are lying in the background. Similarly, in the Tent-Struggle examples, the model focuses on the tent's poles and the hands of the person attempting to put it up, as shown by the heatmaps. In the Tower-Struggle examples, the heatmaps mainly lie on the hands and the Hanoi puzzle blocks. The corresponding green bars below show that the model is giving each frame varying amounts of attention. Since the analysed video clips are limited to 10 seconds and do not capture the full action sequence, the green bars may not align perfectly with explicit actions. However, they still indicate the model's preference for specific frames within each sequence.

\section{Limitations}
\label{sec:limitations}
We acknowledge several limitations and areas for improvement as follows:
(1) Our dataset primarily consists of two indoor environments for pipe assembly and the Tower of Hanoi game, along with one outdoor setting for tent assembly. This may limit its generalizability to more diverse backgrounds and is also constrained by the amount of available data.
(2) Struggle in our dataset is annotated based on 10-second trimmed video segments, which may limit the ability to achieve finer-grained temporal localization and capture struggle over longer temporal dependencies.

\section{Discussion and Conclusions}
\label{sec:discussion}
In this paper, we introduced the struggle determination task as an important component in video understanding. 
We collected a new dataset for struggle determination, comprising three daily activities--Pipes-Struggle, Tent-Struggle, and Tower-Struggle--with a total of 5.09 hours of video recordings. Struggle annotations were provided at the video level by both human experts and crowd voters. 

Our experiments evaluated various deep learning models as baselines for struggle determination. Among the different modelling approaches, struggle classification proved to be the most effective. The struggle classification results reveal opportunities for improving four-way struggle classification, which particularly focuses on finer-grained struggle assessment. The similar accuracy between four-way to binary conversion and binary classification suggests that the main challenge lies in distinguishing different degrees of struggle—an essential factor for developing reliable assistive systems. 

We further conducted struggle generalization experiments across the activities showcasing the challenges, and we also compared different models to identify the most efficient backbone model architectures for struggle determination. Our findings in the ablation study highlight the importance of temporal motion information and the correct frame sequence for struggle determination. Additionally, we report baseline results for alternative training strategies, treating struggle determination as regression and label distribution learning tasks.

In conclusion, our research work in this paper lays the foundation for further research in struggle determination from videos. We hope this research provides valuable insights for additional work on fine-grained egocentric video understanding and supports new methods and directions to benchmark assistive systems. 

\backmatter

\bmhead{Dataset Access}
Our struggle determination dataset can be found via this URL: 
\noindent
\url{https://github.com/FELIXFENG2019/Struggle-Determination}.

\bmhead{Acknowledgments}
We are grateful to Prof. Dima Damen's contributions to early discussions and involvement in the initial investigation of this work. Shijia Feng is supported by a scholarship from the China Scholarship Council (No.202109210007). Initial research, data capture and annotation supported by EPSRC grant GLANCE (EP/N013964/1).










\begin{appendices}





\section{Related Datasets}
\label{related datasets}

\begin{sidewaystable*}[htp]
\tiny
\centering
\captionof{table}{\textbf{Comparison of related datasets.} Note that AR is short for Action Recognition. AQA: Action Quality Assessment. VAD: Video Anomaly Detection.}
\label{tab:related-datasets}
\rotatebox{0}{%
\begin{tabularx}{\textwidth}{
    >{\hsize=.12\hsize}X
    >{\hsize=.03\hsize}X
    >{\hsize=.11\hsize}X
    >{\hsize=.08\hsize}X
    >{\hsize=.26\hsize}X
    >{\hsize=.26\hsize}X
    >{\hsize=.14\hsize}X
}
\toprule
Dataset & Year & Source & Viewpoint & Tasks & Available Annotations & Size \\
\midrule
\multicolumn{7}{l}{\it{Coarse-grained AR}} \\ \cmidrule{1-1}
HMDB51 \citep{Kuehne11} & 2011 & Movies and online & Third-person & Human motion recognition & 51 action categories & 6,766 video clips \\ \cmidrule{2-7}
UCF101 \citep{soomro2012ucf101} & 2012 & YouTube & Third-person & Action recognition & 101 action classes & Over 13k clips \\ \cmidrule{2-7}
ActivityNet \citep{caba2015activitynet} & 2015 & YouTube & Third-person & Human activity understanding & 200 activities & 19,994 untrimmed videos \\ \cmidrule{2-7}
Kinetics \citep{https://doi.org/10.48550/arxiv.1705.06950} & 2017-2020 & YouTube & Third-person & Action recognition & 400/600/700 human action classes & Up to 650,000 video clips \\ \midrule
\multicolumn{7}{l}{\it{Fine-grained AR}}  \\ \cmidrule{1-1}
SSV2 \citep{https://doi.org/10.48550/arxiv.1706.04261} & 2017 & In-person recording & First-person & Fine-grained understanding of human-object interaction & 174 labels for different types of hand-object interactions & 108,499 videos \\ \cmidrule{2-7}
Diving48 \citep{li2018resound} & 2018 & Web videos & Third-person & Fine-grained diving action recognition & 48 dive classes & About 18k videos \\ \cmidrule{2-7}
FineGym \citep{shao2020finegym} & 2020 & YouTube & Third-person & Fine-grained action recognition & 15 set classes, 530 element classes, 4,883 instances, 32,697 sub-instances &  303 competition records \\ \midrule
\multicolumn{7}{l}{\it{AQA}}        \\ \cmidrule{1-1}
Olympic Scoring \citep{https://doi.org/10.48550/arxiv.1611.05125} & 2017 & Existing datasets and online & Third-person & Olympics sports score assessment & score ranges from 0 or 20 to 100 & 696 videos \\ \cmidrule{2-7}
JIGSAW \citep{gao2014jhu} & 2017 & JHU and Sunnyvale, CA. ISI & First-person & Surgery skill assessment & 15 gestures and 5-point Likert scale rating score & 206 videos \\ \cmidrule{2-7}
AQA-7 \citep{parmar2019action} & 2019 & YouTube & Third-person & Action quality assessment across seven sports & AQA scores range from 6.72 to 104.88 altogether across seven sports & 1189 videos \\ \cmidrule{2-7}
MTL-AQA \citep{mtlaqa} & 2019 & Online & Third-person & Multi-task learning including fine-grained action recognition, commentary generation, AQA score estimation & 16 classes of events, 5,748 commentaries, and AQA scores range from 0 to 100 & 1,412 video samples \\ \cmidrule{2-7}
FineDiving \citep{https://doi.org/10.48550/arxiv.2204.03646} & 2022 & YouTube  & Third-person & Coarse- to fine-grained diving action procedures and AQA & 52 action types and 29 sub-action types with temporal boundaries, along with judges’ scores, and 23 difficulty degree types. & 3,000 video samples \\ \midrule
\multicolumn{7}{l}{\it{Skill Determination}}                \\ \cmidrule{1-1}
EPIC-Skill2018 \citep{Doughty_2017_CVPR}  & 2018 & Existing dataset and in-person recording & First-person & Pairwise skill ranking including four tasks & Lists of video 1, video 2, and the better video in the pair. & 196 video samples \\ \cmidrule{2-7}
BEST \citep{Doughty_2019_CVPR}  & 2019 & In-person recording & First-person & Pairwise skill ranking including five tasks & Lists of video 1, video 2, and the better video in the pair. & 500 videos \\  \cmidrule{2-7}
TikTok Dance \citep{Whodance2023}  & 2023 & Participants on TikTok & Third-person & Pairwise dance skill ranking tasks including 12 dance challenges & Video pairs with the winning (better) video. & 240 videos \\ \midrule
\multicolumn{7}{l}{\it{VAD}}          \\ \cmidrule{1-1}
UCSD-Peds \citep{mahadevan2010anomaly}  & 2013 & Surveillance camera & Third-person & Determine videos with abnormal actions & Binary per-frame flag indicator and pixel-level binary masks for abnormal regions & 70 videos in Ped 1 and 28 videos in Ped 2 \\ \cmidrule{2-7}
ShanghaiTech \citep{liu2018ano_pred} & 2018 & Surveillance camera & Third-person & Predict future frames and determine abnormal events & Frame-level masks and pixel-level masks for abnormal events & 437 videos \\ \cmidrule{2-7}
UCF-Crime \citep{Sultani_2018_CVPR}  & 2018 & Surveillance videos & Third-person & Determine videos with abnormal actions & Video-level weak labels for anomalies & 1,900 untrimmed videos  \\ \midrule
\multicolumn{7}{l}{\it{Struggle Determination}}          \\ \cmidrule{1-1}
\textbf{Ours}  & 2023 & Existing dataset and in-person recording & First-person & Determine struggle and degrees of struggle under three sub-tasks: struggle classification, struggle level regression, and struggle label distribution learning & Video-level binary labels for struggle/non-struggle or four-scale struggle levels from 1 to 4. & 1,832 trimmed video segments. \\ 
\bottomrule
\end{tabularx}%
}
\end{sidewaystable*}

Table~\ref{tab:related-datasets} contains a comparison of datasets in video understanding previously used for coarse-grained Action Recognition (AR), fine-grained AR, Action Quality Assessment (AQA), Skill Determination, Video Anomaly Detection (VAD) and Struggle Determination. These datasets are relevant to our struggle determination task as discussed in Section \ref{sec: related work}. 

Action recognition datasets encompass both coarse-grained and fine-grained human actions. Deep models trained on large-scale action recognition datasets can be transferred to various downstream tasks due to the diverse scenarios and learned features. However, these datasets do not focus on scenes depicting struggling, particularly in the case of coarse-grained AR, which mainly includes common human actions and lacks scenarios for skills.

Action quality assessment datasets primarily annotate action quality scores for videos showcasing sports or surgical skills. However, lower action quality scores do not necessarily indicate struggling, and well-performed actions can still receive relatively low scores.

Skill determination datasets are more relevant to our struggle determination task as they often showcase different skills at different levels with hand-object interactions. However, the annotations in these datasets are in the form of pairwise rankings, indicating which skill is better than the other. Lower-ranked videos in these pairs may not guarantee that a person is struggling. 

Video anomaly detection datasets mainly consist of surveillance videos capturing unusual events, such as fights or unintentional falls. These datasets are typically used for binary classification tasks to distinguish between anomaly and normal videos. While struggle determination also involves binary classification, it falls within the context of determining skills. Moreover, our goal is not only to differentiate struggle from non-struggle but also to determine the degree of struggle, which is crucial for developing a wearable intelligent system to provide appropriate assistance to the user. 

In comparison, our struggle determination datasets encompass three tasks that demonstrate different types of skills in assembling pipes, pitching tents, and playing the Tower of Hanoi game. These datasets cover indoor and outdoor scenes and include both rigid and non-rigid objects. Additionally, we specifically use a struggle scoring scale. Thus, the annotations in our dataset are not limited to binary labels but are further fine-grained to indicate the degree of struggle. Aside from the focus on struggle, this distinguishes them from the class labels in AR, action quality scores in AQA, ranking labels in skill determination, and even the binary contrastive labels in VAD. Hence, it is essential to create a dataset with a focus on struggle annotations to train deep models for determining struggle and quantifying its degree.

\section{Dataset Collection Details}

\subsection{Instructions for Participants}
\label{instructions}

\begin{figure*}[ht]
  \centering
  \includegraphics[width=0.7\linewidth]{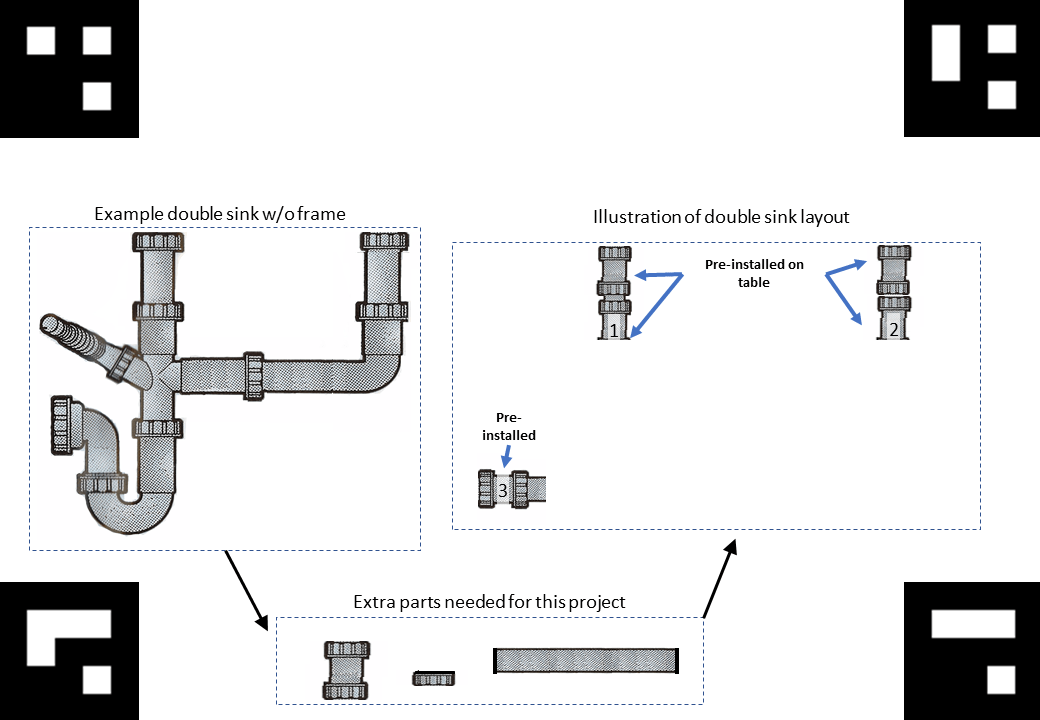}
  \caption{\textbf{The diagram of the \emph{easy} double sink layout.}}
  \label{fig:pipes-ins-easy}
\end{figure*}

\begin{figure*}[ht]
  \centering
  \includegraphics[width=0.7\linewidth]{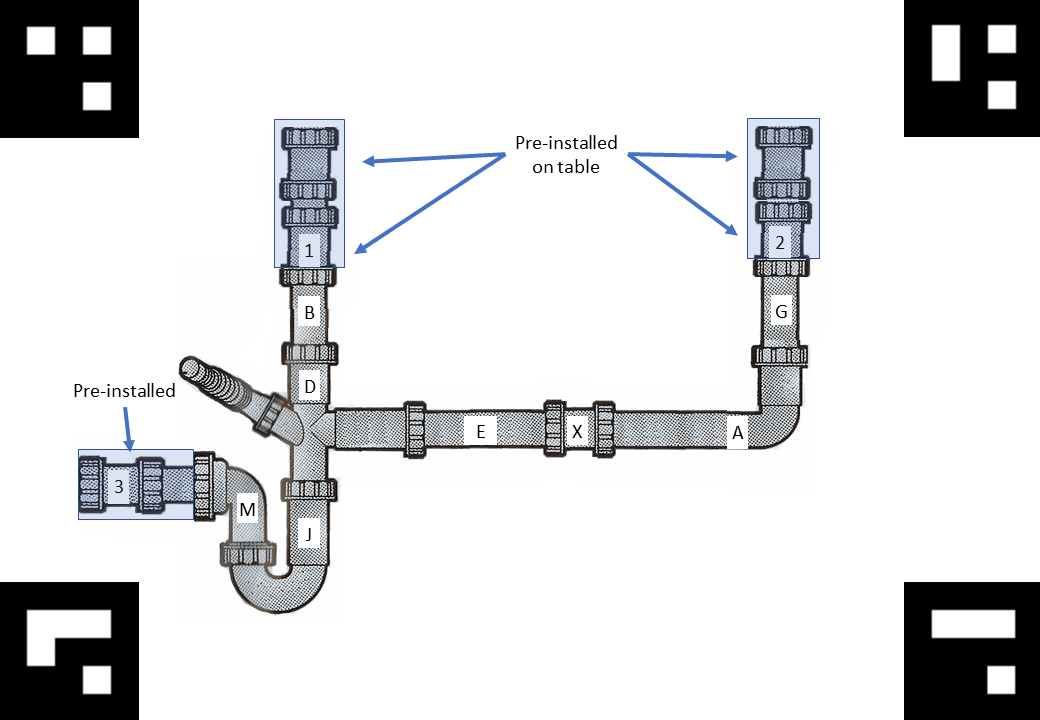}
  \caption{\textbf{The diagram of the \emph{difficult} double sink layout.}}
  \label{fig:pipes-ins-hard}
\end{figure*}

\begin{figure*}[ht]
  \centering
  \includegraphics[width=0.9\linewidth]{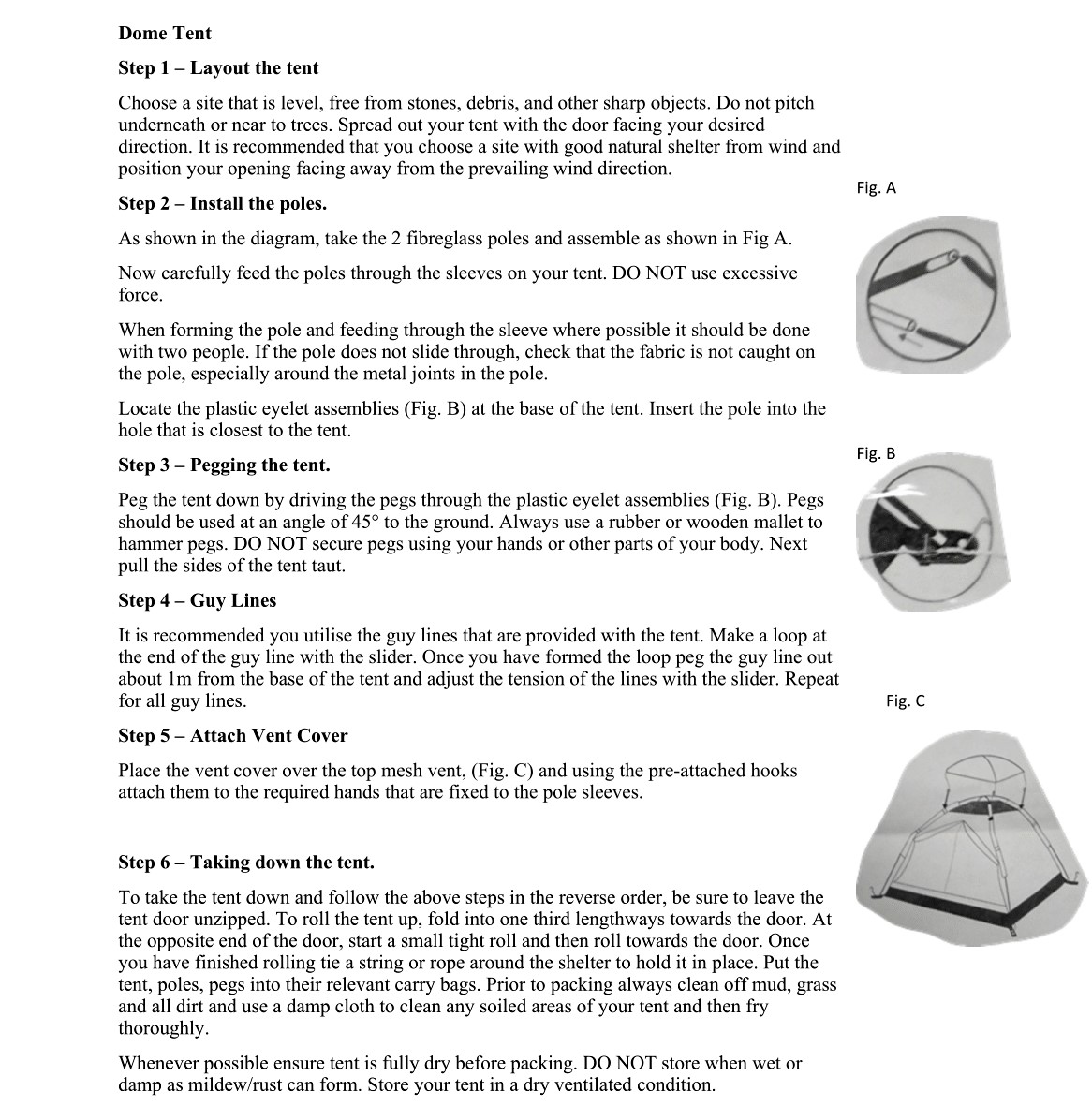}
  \caption{\textbf{An OCR photocopy of the assembly instructions for the tent we use.}}
  \label{fig:tent-instructions}
\end{figure*}

During our data collection process, instructions were given to our in-person participants both verbally and written on paper. For the plumbing pipes task, there are two instruction diagrams illustrating how to assemble the pipes correctly. One version is easy to follow, shown in Figure~\ref{fig:pipes-ins-easy}, and another more difficult version, shown in Figure~\ref{fig:pipes-ins-hard} so that the participants using a hard version of instructions are more likely to show signs of struggle. For the tent-pitching task, we used the paper instructions from the manufacturer, as shown in Figure~\ref{fig:tent-instructions}. For the Tower of Hanoi game task, the researchers only provided the instructions verbally introducing the basic rules of the game, such as only one disk can be moved at a time, and a larger disk cannot be placed on top of a smaller one.  

\subsection{Additional Dataset Statistics}
\label{more statistics}

\begin{figure*}[ht]
    \centering
    \includegraphics[width=0.9\linewidth]{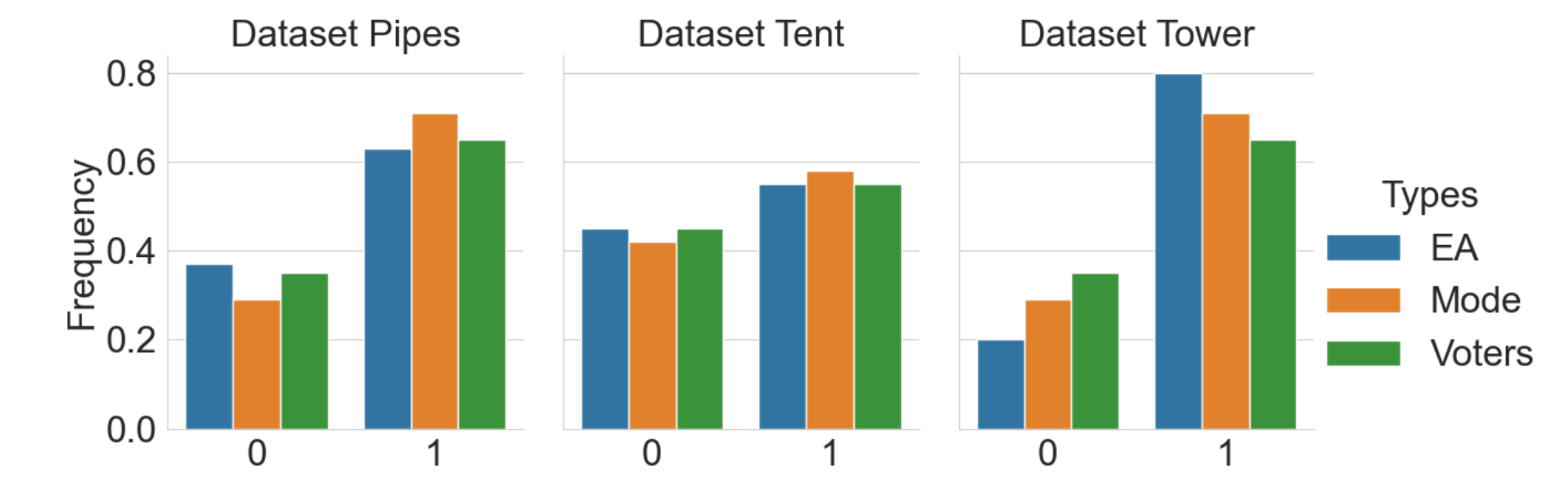}
    \caption{\textbf{Comparison of the distributions using binary struggle labels.} The three graphs display the frequency distribution of the binary struggle labels (0 for non-struggle, 1 for struggle) including \textcolor{blue}{expert annotation (EA)}, \textcolor{orange}{voters mode (Mode)}, and \textcolor{green}{voters labels distributions (Voters)}, for the three activities: Pipes-Struggle, Tent-Struggle, and Tower-Struggle, respectively.}
    \label{fig-appendix:datasets-stats-binary}
\end{figure*}

We add Figure~\ref{fig-appendix:datasets-stats-binary} to display the struggle labels frequency distributions in the binary case (non-struggle vs struggle). 

To analyse intra-group voter disagreements, we converted the four-way struggle labels into binary labels and examined the distribution of disagreement levels among voters. The disagreement frequency for each video sample was computed by determining the proportion of voters who selected either struggle or non-struggle, relative to the total number of voters. The smaller of these two proportions was designated as the minority group frequency, representing the level of disagreement among voters.

\begin{figure*}[ht]
    \centering
    \includegraphics[width=1.0\linewidth]{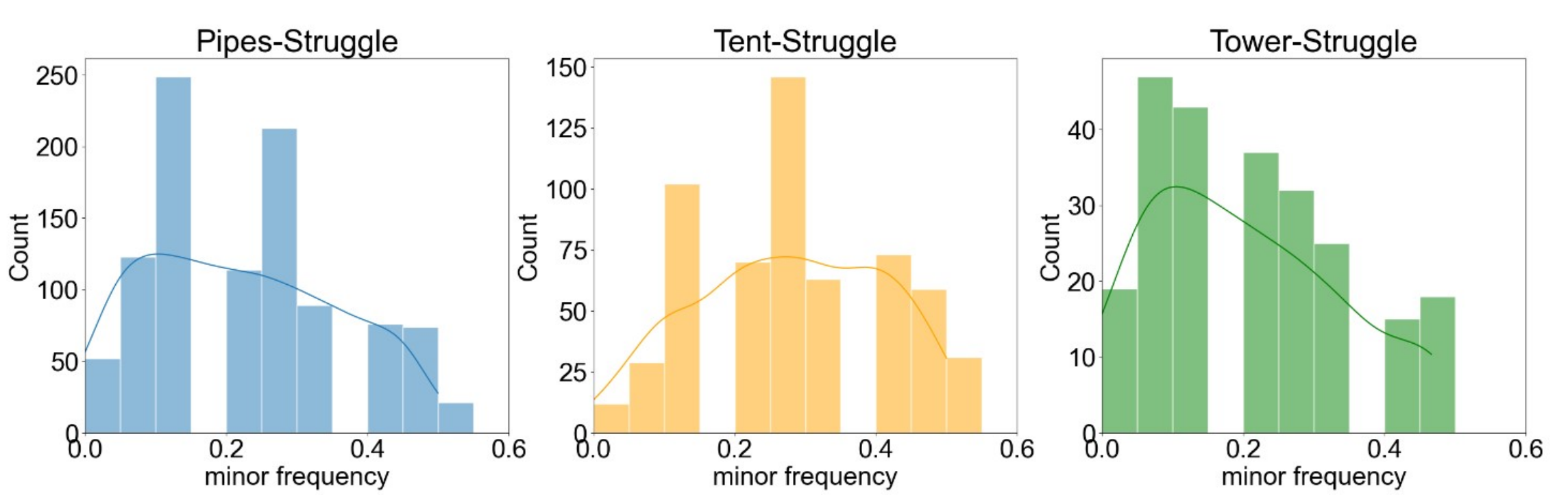}
    \caption{\textbf{Intra-group voter disagreements based on the binary struggle labels.} The three graphs illustrate the frequency distribution of disagreements after converting the four-way struggle labels into binary labels. Each graph presents a histogram of video sample counts (y-axis) against the frequency of the minority group of voters (x-axis), with a smoothed kernel density estimate (KDE) curve.}
    \label{fig-appendix:struggle_intragroup_disagreement}
\end{figure*}

We show the intra-group disagreement in Figure~\ref{fig-appendix:struggle_intragroup_disagreement} using three histograms to illustrate the distribution of these minority group frequencies across the three activities. The x-axis represents the frequency of the minority group of voters, ranging from 0 to 0.5, while the y-axis indicates the number of video samples. Additionally, a smoothed KDE curve is overlaid to highlight the density of disagreements within the dataset.

\section{Additional Experiment Results}
\label{addtional-experiment-results}
\subsection{Struggle Modelling Approach Comparison}
\label{add-cls-reg-comparison}
We also conducted extensive experiments comparing classification and regression results across different deep models to assess the consistency of this trend (see Table~\ref{tab:tsn_comparison} and \ref{tab:slowfast_comparison}). The results show that for the TSN~\citep{wang2016temporal} model, there are three instances where classification achieves higher four-way accuracy (Pipes-Struggle and Tower-Struggle) and binary accuracy (Pipes-Struggle), while regression-to-classification slightly outperforms classification in some cases (e.g., binary accuracy for Tower-Struggle and four-way accuracy for Tent-Struggle). However, classification remains the more stable and effective approach overall. Similarly, for the SlowFast-R50~\citep{feichtenhofer2019slowfast} model, classification consistently outperforms regression-to-classification across all activities in binary accuracy and, in most cases, for four-way accuracy. These findings consistently indicate that models trained for classification achieve superior performance compared to those trained for regression, reinforcing the advantage of direct classification training for struggle detection.

\begin{table*}[ht]
\scriptsize
\centering
\caption{Comparison of Classification and Regression-to-Classification Accuracy for TSN~\citep{wang2016temporal}.}
\label{tab:tsn_comparison}
\begin{tabular}{lcccc}
\toprule
\multirow{2}{*}{\textbf{Activity}} & \multicolumn{2}{c}{\textbf{Classification}} & \multicolumn{2}{c}{\textbf{Regression-to-Classification}} \\
\cmidrule(lr){2-3} \cmidrule(lr){4-5}
                  & Binary Accuracy & Four-Way Accuracy & Binary Accuracy & Four-Way Accuracy \\
\midrule
Pipes-Struggle    & \textbf{74.54}         & \textbf{48.59}            & 72.42         & 45.02 \\
Tent-Struggle     & \textbf{68.05}         & 38.23            & \textbf{68.27}         & \textbf{40.95} \\
Tower-Struggle    & 82.59         & \textbf{59.49}            & \textbf{85.35}         & 58.01 \\
\bottomrule
\end{tabular}
\end{table*}

\begin{table*}[ht]
\scriptsize
\centering
\caption{Comparison of Classification and Regression-to-Classification Accuracy for SlowFast-R50~\citep{feichtenhofer2019slowfast}.}
\label{tab:slowfast_comparison}
\begin{tabular}{lcccc}
\toprule
\multirow{2}{*}{\textbf{Activity}} & \multicolumn{2}{c}{\textbf{Classification}} & \multicolumn{2}{c}{\textbf{Regression-to-Classification}} \\
\cmidrule(lr){2-3} \cmidrule(lr){4-5}
                  & Binary Accuracy & Four-Way Accuracy & Binary Accuracy & Four-Way Accuracy \\
\midrule
Pipes-Struggle    & \textbf{78.01}         & \textbf{50.77}            & 76.89         & 45.28 \\
Tent-Struggle     & \textbf{69.56}         & 39.48            & \textbf{69.06}         & \textbf{41.63} \\
Tower-Struggle    & \textbf{88.24}         & \textbf{63.92}            & 86.89         & 56.00 \\
\bottomrule
\end{tabular}
\end{table*}

\subsection{Struggle Classification}
\label{add-cls}

We show more details of our struggle classification modelling approach about the top-1 accuracy rate for each of the four splits together with the mean and standard deviation (StdDev) over the accuracy of the four splits. 
Table \ref{tab:full cls bin}, \ref{tab:full cls fourway}, and \ref{tab:full cls 4to2} show the additional results for binary classification, four-way classification, and four-way to binary classification, respectively. 

We first give the full definitions of the three naive baselines: random, majority class, and voters' baseline. Given a dataset $D$ (Pipes-Struggle, Tent-Struggle, or Tower-Struggle), $N$ is the number of videos. Let $V_i$ denote the $i$th video. The expert annotation for video $V_i$ is $y^e_i$ and the frequency distribution of the struggle labels given by the 15 or 20 voters $Dist_i$ is denoted as $Dist_i=\{f_1, f_2, f_3, f_4\}$ where $f_1$ is the frequency of the voters label for struggle level 1 (definitely non-struggle), noted as $Dist_i(f_1)$; $f_2$ is the frequency of the voters label for struggle level 2 (slightly non-struggle), noted as $Dist_i(f_2)$; $f_3$ is the frequency of the voters label for struggle level 3 (slightly struggle), noted as $Dist_i(f_3)$; $f_4$ is the frequency of the voters label for struggle level 4 (definitely struggle), noted as $Dist_i(f_4)$, while $f_1+f_2+f_3+f_4=1$. Let the mode of the voters' label for video $V_i$ be $y^l_i$. $y^l_i$ is the struggle label index (from 1 to 4 ) of $Mode(Dist_i)$. Note that in the case of binary classification, we combine 1 and 2 to 0 (non-struggle), and 3 and 4 to 1 (struggle) as discussed in the main paper. Moreover, the label distributions over a certain split are calculated by averaging the expert annotation $y^e_i$ or the voters' label distributions $Dist_i=\{f_1, f_2, f_3, f_4\}$ across all the videos $V_i$ within the split, denoted as $Dist(y_0)$ and $Dist(y_1)$ respectively.  

In this way, for the random baseline, we assume the random model only has the ability to predict each of the struggle levels in a probability that corresponds to the proportions of the expert annotations in the test split $Dist(y_0)$, so that we can calculate the top-1 accuracy rate each time we run the random model. When the number of runs tends to be a large number close to infinity, the averaged top-1 accuracy rate over the number of runs is approaching a set percentage which is the random baseline accuracy. The majority class baseline is the percentage of the maximum entity in $Dist(y_0)$ for a split. Finally, we illustrate the voter's baseline. Given a video $V_i$, we define the Kronecker delta function:
\begin{equation}
    \delta_{y_{i}^{e}y_{i}^{l}}=
    \begin{cases} 
        1 & \text{if } y_{i}^{e}=y_{i}^{l} \\ 
        0 & \text{if } y_{i}^{e}\ne y_{i}^{l} 
    \end{cases}
\end{equation}
where $y_{i}^{e}$ is the expert annotation and $y_{i}^{l}$ is the mode of the voters' labels. So the voters' baseline accuracy is calculated as:
\begin{equation}
    \frac{1}{N_j}\sum_{i=1}^{N_j}\delta_{y_{i}^{e}y_{i}^{l}}\times 100\%
\end{equation}
where the $N_j$ is the number of videos in a split $j$. 

\subsection{Struggle VLM Baselines}
\label{add-vlm}

\begin{table*}[ht]
\caption{\textbf{Struggle classification results with VLM baseline.} Classification accuracy for binary classification, four-way classification and the four-way to binary classification. 
}
\label{table:classification results with vlm}
\centering
\resizebox{\textwidth}{!}{%
\begin{tabular}{@{}lccccccccccc@{}}
\toprule
\multirow{2}{*}{Dataset}                                   & \multicolumn{3}{c}{Pipes-Struggle}                                                               & \multicolumn{1}{c}{} & \multicolumn{3}{c}{Tent-Struggle}                                                           & \multicolumn{1}{c}{} & \multicolumn{3}{c}{Tower-Struggle}                                                          \\ 
                                           & \multicolumn{3}{c}{Top-1 Accuracy (\%)}                                                          & \multicolumn{1}{c}{} & \multicolumn{3}{c}{Top-1 Accuracy (\%)}                                                     & \multicolumn{1}{c}{} & \multicolumn{3}{c}{Top-1 Accuracy (\%)}                                                     \\ \cmidrule(lr){2-4} \cmidrule(lr){6-8} \cmidrule(l){10-12} 
Models                                     & \multicolumn{1}{c}{Binary} & \multicolumn{1}{c}{Four-Way} & \multicolumn{1}{c}{Four-to-Bin} & \multicolumn{1}{c}{} & \multicolumn{1}{c}{Binary} & \multicolumn{1}{c}{Four-Way} & \multicolumn{1}{c}{Four-to-Bin} & \multicolumn{1}{c}{} & \multicolumn{1}{c}{Binary} & \multicolumn{1}{c}{Four-Way} & \multicolumn{1}{c}{Four-to-Bin} \\ \midrule
Random                                     & 53.38                        & 28.62                           & 53.38                           &                      & 50.59                      & 27.12                        & 50.59                           &                      & 67.86                      & 40.45                        & 67.86                           \\
Majority Class                             & 62.60                        & 39.60                           & 62.60                           &                      & 54.91                      & 35.79                        & 54.91                           &                      & 79.74                      & 57.56                        & 79.74                           \\
Voters' Baseline                             & 80.56                        & 45.31                           & 80.56                           &                      & 72.63                      & 43.12                        & 72.63                           &                      & 84.41                      & 51.86                        & 84.41                           \\ 
Video-LLaVA~\citep{lin2023video} & 37.40 & 14.45 & 37.40 & & 48.03 & 24.10 & 48.03 & & 20.26 & 12.12 & 20.26 \\ \bottomrule
\end{tabular}%
}
\end{table*}

We used pre-trained Video-LLaVA~\citep{lin2023video} and evaluated the model’s struggle classification performance on our datasets in a zero-shot setting without further fine-tuning. The evaluation results are shown in Table~\ref{table:classification results with vlm}. 

We can see that the Video-LLaVA~\citep{lin2023video} VLM model does not perform very well in both binary and four-way struggle classification. The highest binary classification accuracy is still lower than the random level (48.03 vs 53.38), and the situation remains the same on four-way struggle classification, indicating that the Video-LLaVA~\citep{lin2023video} model struggles to capture struggle-specific visual cues given text prompts and video input. This is reasonable because the VLM model is not trained or fine-tuned specifically on struggle determination tasks, and the struggle is distinct from many video-understanding tasks on which the VLM model has been trained. 

We designed structured input prompts to guide the VLM in identifying struggles. Below is an example of the input prompt used for the Tent-Struggle dataset: ``USER: This video shows a person pitching a camping tent in a random 10-second temporal window. Please focus on the hand's movements and object status and pay attention to the visual signs that may indicate a struggle, such as motor hesitation (e.g., stopping or placement indecision), getting stuck, prolonged actions, non-smooth movements, frequent pauses, repeated attempts, or visible signs of frustration (e.g., hand or head movements). Describe what you observe before deciding if the person is struggling or not. ASSISTANT: "

To further prompt the VLM to classify struggle and struggle level, we add the following text input: ``If no struggle indicators are present, state `None observed.' Then, classify the struggle level on a scale from 1 to 4: (1) definitely non-struggle – smooth, confident movements; (2) slightly non-struggle – minor hesitation but controlled actions; (3) slightly struggle – noticeable pauses, reattempts, or mild frustration; (4) definitely struggle – frequent failed attempts, prolonged hesitation, or strong frustration cues. Provide the output in this format: Observation: [description] Struggle Indicators: [if any] Struggle Level: [1, 2, 3, or 4].”

The text output from the Video-LLaVA~\citep{lin2023video} VLM model reveals that it suffers from hallucination when handling unseen struggle-related data. In many cases, the Assistant's responses describe events that are not present in the video. This issue likely arises from the model's tendency to rely more on textual priors than actual visual input. Furthermore, because struggle-related tasks are not explicitly represented in the VLM’s training data and differ from conventional visual tasks, the model may struggle to extract the necessary temporal features for accurately detecting struggle.

\subsection{Struggle Regression}
\label{add-reg}

We provide more experimental results on our struggle level regression modelling approach with the evaluation metrics of Mean Squared Error (MSE), Mean Absolute Error (MAE), and the top-1 accuracy rate of binary and four-way classification converted from the predicted struggle level regression scores. These evaluation metrics of various deep models on the three activities for split 1 to 4 are shown in Tables \ref{tab:reg split 1}, \ref{tab:reg split 2}, \ref{tab:reg split 3}, and \ref{tab:reg split 4}, respectively. The mean and standard deviation (StdDev) for these evaluation metrics for the four splits are shown in Table \ref{tab:reg stats mean-std}.  

\subsection{Struggle Label Distribution Learning}
\label{add-ldl}

We provide more experimental results about voters' struggle label distribution learning modelling approach with the evaluation metrics of Mean Absolute Error (MAE), Spearman's Rank Correlation (Spearman's Rho), and the top-1 accuracy rate of binary and four-way classification by selecting the struggle level classes with the highest frequency in the label distributions and using the voters' mode label as ground truth. These evaluation metrics of various deep models on the three activities for split 1 to 4 are shown in Tables \ref{tab:ldl split 1}, \ref{tab:ldl split 2}, \ref{tab:ldl split 3}, and \ref{tab:ldl split 4}, respectively. The mean and standard deviation (StdDev) for these evaluation metrics for the four splits are shown in Table \ref{tab:ldl stats mean stddev}.

\subsection{Multiple Frames and Temporal Ordering}
\label{sec: ablation table results}

We provide more experiment results of the ablation study we conducted using one frame as input and shuffled frame orders, see Table \ref{table:one frame} and Table \ref{table:shuffle frame}. 

\section{Figures and Tables}
\label{sec: figs and tabs}

\clearpage

\clearpage

\begin{table*}[ht]
\centering
\caption{\textbf{Experiment results for struggle binary classification over all splits.}}
\label{tab:full cls bin}
\resizebox{\textwidth}{!}{
\begin{tabular}{c|l|ccccc}
\toprule
\multirow{2}{*}{Activities} & \multirow{2}{*}{Models} & \multicolumn{5}{c}{Top1 Accuracy Rate (\%)   - Binary Classification} \\
                            &                  & Split 1 & Split 2 & Split 3 & Split 4 & Mean\textpm StdDev \\
\midrule
\multirow{7}{*}{EPIC-Pipes} & Random     & 55.00\% & 51.60\% & 51.93\% & 54.98\% & 53.38\textpm1.87\% \\
                            & Majority Class   & 65.82\% & 58.95\% & 59.84\% & 65.78\% & 62.60\textpm3.72\% \\
                            & Voters' Baseline   & 83.27\% & 79.91\% & 80.33\% & 78.71\% & 80.56\textpm1.68\% \\ \cmidrule{2-7}
                            & TSN \citep{wang2016temporal}  & 75.27\% & \textbf{76.86\%} & 74.18\% & 71.86\% & 74.54\textpm2.10\% \\
                            & SlowFast-R50 \citep{feichtenhofer2019slowfast}  & 77.09\% & 75.11\% & \textbf{80.74\%} & \textbf{79.09\%} & \textbf{78.01\textpm2.44\%} \\
                            & SlowFast-gMLP    & \textbf{78.91\%} & \textbf{76.42\%} & \textbf{80.33\%} & 78.71\% & \textbf{78.59\textpm1.62\%} \\
                            & VTN-SlowFast-ViT & 77.82\% & 75.55\% & 79.51\% & 78.33\% & 77.80\textpm1.66\% \\
                            & MViTv2 \citep{li2022mvitv2} & 76.73\% & 73.80\% & 77.46\% & 78.71\% & 76.68\textpm2.08\% \\
                            \midrule
\multirow{7}{*}{EPIC-Tent}  & Random     & 50.37\% & 51.02\% & 50.94\% & 50.03\% & 50.59\textpm0.47\% \\
                            & Majority Class   & 54.35\% & 57.14\% & 56.85\% & 51.30\% & 54.91\textpm2.71\% \\
                            & Voters' Baseline   & 79.71\% & 73.47\% & 68.49\% & 68.83\% & 72.63\textpm4.54\% \\ \cmidrule{2-7}
                            & TSN \citep{wang2016temporal}    & 74.64\% & 70.07\% & \textbf{66.44\%} & 61.04\% & 68.05\textpm5.75\% \\
                            & SlowFast-R50  \citep{feichtenhofer2019slowfast}   & \textbf{76.81\%} & 72.11\% & 64.38\% & \textbf{64.94\%} & \textbf{69.56\textpm5.98\%} \\
                            & SlowFast-gMLP    & 73.19\% & \textbf{73.47\%} & 63.01\% & 62.34\% & 68.00\textpm6.16\% \\
                            & VTN-SlowFast-ViT & 73.91\% & 72.79\% & 65.07\% & 62.34\% & 68.53\textpm5.70\% \\
                            & MViTv2 \citep{li2022mvitv2} & 68.12\% & 68.03\% & 64.38\% & 57.14\% & 64.42\textpm5.16\% \\
                            \midrule
\multirow{7}{*}{EPIC-Tower} & Random     & 71.88\% & 65.43\% & 70.59\% & 63.53\% & 67.86\textpm4.01\% \\
                            & Majority Class   & 83.08\% & 77.78\% & 82.09\% & 76.00\% & 79.74\textpm3.39\% \\
                            & Voters' Baseline   & 81.54\% & 87.04\% & 85.07\% & 84.00\% & 84.41\textpm1.98\% \\ \cmidrule{2-7}
                            & TSN \citep{wang2016temporal} & 86.15\% & 79.63\% & 86.57\% & 78.00\% & 82.59\textpm4.41\% \\
                            & SlowFast-R50  \citep{feichtenhofer2019slowfast}   & \textbf{89.23\%} & \textbf{85.18\%} & \textbf{92.54\%} & \textbf{86.00\%} & \textbf{88.24\textpm3.36\%} \\
                            & SlowFast-gMLP    & \textbf{89.23\%} & 85.18\% & 91.04\% & \textbf{86.00\%} & 87.86\textpm2.75\% \\
                            & VTN-SlowFast-ViT & \textbf{89.23\%} & 85.18\% & 94.03\% & 84.00\% & \textbf{88.11\textpm4.54\%} \\
                            & MViTv2 \citep{li2022mvitv2} & 84.62\% & \textbf{94.44\%} & 82.09\% & 76.00\% & 84.29\textpm7.67\% \\
\bottomrule
\end{tabular}%
}
\end{table*}

\begin{table*}[ht]
\centering
\caption{\textbf{Experiment results for struggle four-way classification over all splits.}}
\label{tab:full cls fourway}
\resizebox{\textwidth}{!}{
\begin{tabular}{c|l|ccccc}
\toprule
\multirow{2}{*}{Activities} & \multirow{2}{*}{Models} & \multicolumn{5}{c}{Top1 Accuracy Rate (\%)   - Four-way Classification} \\
                            &                  & Split 1 & Split 2 & Split 3 & Split 4 & Mean\textpm StdDev \\
                            \midrule
\multirow{7}{*}{EPIC-Pipes} & Random     & 28.82\% & 29.65\% & 27.88\% & 28.12\% & 28.62\textpm0.69\% \\
                            & Majority Class   & 40.36\% & 41.92\% & 38.11\% & 38.02\% & 39.60\textpm1.63\% \\
                            & Voters' Baseline   & 47.27\% & 46.29\% & 39.75\% & 47.91\% & 45.31\textpm3.26\% \\ \cmidrule{2-7}
                            & TSN \citep{wang2016temporal}   & 48.00\% & \textbf{53.71\%} & 45.49\% & 47.15\% & 48.59\textpm3.09\% \\
                            & SlowFast-R50 \citep{feichtenhofer2019slowfast}    & \textbf{53.46\%} & 50.66\% & 48.77\% & 50.19\% & \textbf{50.77\textpm1.70\%} \\
                            & SlowFast-gMLP    & \textbf{53.46\%} & 48.04\% & 46.72\% & 50.57\% & 49.70\textpm2.58\% \\
                            & VTN-SlowFast-ViT & 50.54\% & 51.53\% & 48.77\% & \textbf{52.47\%} & \textbf{50.83\textpm1.37\%} \\
                            & MViTv2 \citep{li2022mvitv2} & 48.00\% & 47.60\% & \textbf{52.87\%} & 45.25\% & 48.43\textpm3.20\% \\
                            \midrule
\multirow{7}{*}{EPIC-Tent}  & Random     & 26.71\% & 27.15\% & 28.87\% & 25.73\% & 27.12\textpm1.14\% \\
                            & Majority Class   & 35.51\% & 36.05\% & 40.41\% & 31.17\% & 35.79\textpm3.27\% \\
                            & Voters' Baseline   & 50.72\% & 38.10\% & 47.95\% & 35.71\% & 43.12\textpm6.35\% \\ \cmidrule{2-7}
                            & TSN \citep{wang2016temporal}   & 40.58\% & 39.46\% & 40.41\% & 32.47\% & 38.23\textpm3.35\% \\
                            & SlowFast-R50 \citep{feichtenhofer2019slowfast}    & \textbf{45.65\%} & 38.10\% & 39.73\% & 34.42\% & 39.48\textpm4.05\% \\
                            & SlowFast-gMLP    & 41.30\% & \textbf{40.82\%} & 36.99\% & \textbf{38.96\%} & 39.52\textpm1.70\% \\
                            & VTN-SlowFast-ViT & 42.75\% & \textbf{40.14\%} & \textbf{42.47\%} & 35.06\% & \textbf{40.11\textpm3.08\%} \\
                            & MViTv2 \citep{li2022mvitv2} & 38.41\% & 38.78\% & \textbf{42.47\%} & 28.57\% & 37.06\textpm5.95\% \\
                            \midrule
\multirow{7}{*}{EPIC-Tower} & Random     & 42.77\% & 38.54\% & 44.71\% & 35.76\% & 40.45\textpm3.51\% \\
                            & Majority Class   & 60.00\% & 55.56\% & 62.69\% & 52.00\% & 57.56\textpm4.10\% \\
                            & Voters' Baseline   & 56.92\% & 59.26\% & 49.25\% & 42.00\% & 51.86\textpm6.79\% \\ \cmidrule{2-7}
                            & TSN \citep{wang2016temporal} & 63.08\% & 53.70\% & 67.16\% & 54.00\% & 59.49\textpm5.82\% \\
                            & SlowFast-R50 \citep{feichtenhofer2019slowfast}    & 66.15\% & \textbf{64.82\%} & 62.69\% & 62.00\% & 63.92\textpm1.66\% \\
                            & SlowFast-gMLP    & 64.62\% & \textbf{64.82\%} & \textbf{70.15\%} & \textbf{66.00\%} & \textbf{66.40\textpm2.23\%} \\
                            & VTN-SlowFast-ViT & \textbf{67.69\%} & 59.26\% & 67.16\% & 62.00\% & 64.03\textpm3.54\% \\
                            & MViTv2 \citep{li2022mvitv2} & 63.08\% & \textbf{64.82\%} & 47.76\% & 54.00\% & 57.42\textpm8.00\% \\
                            \bottomrule
\end{tabular}%
}
\end{table*}

\begin{table*}[ht]
\centering
\caption{\textbf{Experiment results of four-way classification convert into binary classification over all splits.}}
\label{tab:full cls 4to2}
\resizebox{\textwidth}{!}{
\begin{tabular}{c|l|ccccc}
\toprule
\multirow{2}{*}{Activities} & \multirow{2}{*}{Models} & \multicolumn{5}{c}{Top1 Accuracy Rate (\%)   - Four-way to Binary} \\
                            &                  & Split 1 & Split 2 & Split 3 & Split 4 & Mean\textpm StdDev \\
                            \midrule
\multirow{7}{*}{EPIC-Pipes} & Random     & 55.00\% & 51.60\% & 51.93\% & 54.98\% & 53.38\textpm1.62\% \\
                            & Majority Class   & 65.82\% & 58.95\% & 59.84\% & 65.78\% & 62.60\textpm3.22\% \\
                            & Voters' Baseline   & 83.27\% & 79.91\% & 80.33\% & 78.71\% & 80.56\textpm1.68\% \\ \cmidrule{2-7}
                            & TSN \citep{wang2016temporal} & 73.82\% & 71.52\% & 69.67\% & 70.72\% & 71.43\textpm1.53\% \\
                            & SlowFast-R50 \citep{feichtenhofer2019slowfast} & \textbf{78.18\%} & \textbf{73.36\%} & 74.18\% & \textbf{77.19\%} & \textbf{75.73\textpm2.01\%} \\
                            & SlowFast-gMLP    & \textbf{78.18\%} & 72.93\% & 74.59\% & 72.24\% & 74.49\textpm2.30\% \\
                            & VTN-SlowFast-ViT & 72.73\% & 71.18\% & 73.77\% & 72.24\% & 72.48\textpm0.93\% \\
                            & MViTv2 \citep{li2022mvitv2} & 70.91\% & 68.12\% & \textbf{75.82\%} & 74.14\% & 72.25\textpm3.42\% \\
                            \midrule
\multirow{7}{*}{EPIC-Tent}  & Random     & 50.37\% & 51.02\% & 50.94\% & 50.03\% & 50.59\textpm0.41\% \\
                            & Majority Class   & 54.35\% & 57.14\% & 56.85\% & 51.30\% & 54.91\textpm2.35\% \\
                            & Voters' Baseline   & 79.71\% & 73.47\% & 68.49\% & 68.83\% & 72.63\textpm4.54\% \\ \cmidrule{2-7}
                            & TSN \citep{wang2016temporal} & 67.39\% & 63.95\% & 62.33\% & 51.30\% & 61.24\textpm6.02\% \\
                            & SlowFast-R50 \citep{feichtenhofer2019slowfast} & \textbf{72.46\%} & 70.75\% & \textbf{63.01\%} & 54.55\% & \textbf{65.19\textpm7.10\%} \\
                            & SlowFast-gMLP    & 67.39\% & \textbf{72.11\%} & 57.53\% & \textbf{61.69\%} & 64.68\textpm5.54\% \\
                            & VTN-SlowFast-ViT & 68.84\% & 63.27\% & 55.48\% & 55.84\% & 60.86\textpm5.56\% \\
                            & MViTv2 \citep{li2022mvitv2} & 56.52\% & 61.90\% & 60.27\% & 53.90\% & 58.15\textpm3.62\% \\
                            \midrule
\multirow{7}{*}{EPIC-Tower} & Random     & 71.88\% & 65.43\% & 70.59\% & 63.53\% & 67.86\textpm3.47\% \\
                            & Majority Class   & 83.08\% & 77.78\% & 82.09\% & 76.00\% & 79.74\textpm2.94\% \\
                            & Voters' Baseline   & 81.54\% & 87.04\% & 85.07\% & 84.00\% & 84.41\textpm1.98\% \\ \cmidrule{2-7}
                            & TSN \citep{wang2016temporal} & 83.08\% & 74.07\% & 83.58\% & 74.00\% & 78.68\textpm4.65\% \\
                            & SlowFast-R50 \citep{feichtenhofer2019slowfast} & 84.62\% & 70.37\% & 82.09\% & \textbf{84.00\%} & 80.27\textpm5.79\% \\
                            & SlowFast-gMLP    & \textbf{87.69\%} & 75.93\% & 85.07\% & \textbf{84.00\%} & \textbf{83.17\textpm4.39\%} \\
                            & VTN-SlowFast-ViT & 86.15\% & 75.93\% & \textbf{88.06\%} & 80.00\% & 82.54\textpm4.84\% \\
                            & MViTv2 \citep{li2022mvitv2} & 81.54\% & \textbf{90.74\%} & 82.09\% & 76.00\% & 82.59\textpm6.09\% \\
                            \bottomrule
\end{tabular}%
}
\end{table*}

\begin{table*}[ht]
\scriptsize
\centering
\caption{\textbf{Experiment results for struggle regression---split 1.}}
\label{tab:reg split 1}
\begin{tabular}{c|l|cccc}
\toprule
\multirow{2}{*}{Activities} & \multirow{2}{*}{Models} & \multicolumn{4}{c}{Regression \&   Classification - Split 1} \\
                            &                  & MSE$\downarrow$  & MAE$\downarrow$  & Binary Acc.$\uparrow$ & Four-way Acc.$\uparrow$ \\
                            \midrule
\multirow{3}{*}{EPIC-Pipes} & TSN \citep{wang2016temporal}  & 0.83 & 0.72 & 72.36\%     & 45.82\%       \\
                            & SlowFast-R50 \citep{feichtenhofer2019slowfast} & \textbf{0.64} & 0.67 & \textbf{77.82\%}     & 43.27\%       \\
                            & VTN-SlowFast-ViT & 0.68 & \textbf{0.66} & \textbf{77.45\%}     & \textbf{48.36\%}       \\
                            & MViTv2 \citep{li2022mvitv2} & 0.75 & 0.73 & 75.27\%     & 42.54\%       \\
                            \midrule
\multirow{3}{*}{EPIC-Tent}  & TSN \citep{wang2016temporal}  & 0.82 & 0.74 & 69.57\%     & \textbf{42.03\%}       \\
                            & SlowFast-R50 \citep{feichtenhofer2019slowfast} & \textbf{0.75} & \textbf{0.72} & \textbf{73.91\%}     & \textbf{42.03\%}       \\
                            & VTN-SlowFast-ViT & 0.78 & 0.74 & 68.84\%     & 41.30\%       \\
                            & MViTv2 \citep{li2022mvitv2} & 0.99 & 0.85 & 61.59\%     & 35.51\%       \\
                            \midrule
\multirow{3}{*}{EPIC-Tower} & TSN \citep{wang2016temporal}  & 0.75 & 0.68 & 86.15\%     & 58.46\%       \\
                            & SlowFast-R50 \citep{feichtenhofer2019slowfast} & 0.77 & 0.70  & \textbf{87.69\%}     & 58.46\%       \\
                            & VTN-SlowFast-ViT & \textbf{0.70}  & \textbf{0.67} & \textbf{87.69\%}     & \textbf{60.00\%}       \\
                            & MViTv2 \citep{li2022mvitv2} & 0.82 & 0.78 & 86.15\%     & 30.77\%       \\
                            \bottomrule
\end{tabular}
\end{table*}

\begin{table*}[ht]
\scriptsize
\centering
\caption{\textbf{Experiment results for struggle regression---split 2.}}
\label{tab:reg split 2}
\begin{tabular}{c|l|cccc}
\toprule
\multirow{2}{*}{Activities} & \multirow{2}{*}{Models} & \multicolumn{4}{c}{Regression \&   Classification - Split 2} \\
                            &                  & MSE$\downarrow$  & MAE$\downarrow$  & Binary Acc.$\uparrow$ & Four-way Acc.$\uparrow$ \\
                            \midrule
\multirow{3}{*}{EPIC-Pipes} & TSN \citep{wang2016temporal}  & 0.64 & 0.65 & \textbf{76.42\%}     & \textbf{50.22\%}       \\
                            & SlowFast-R50 \citep{feichtenhofer2019slowfast} & \textbf{0.60}  & \textbf{0.62} & 73.46\%     & 48.47\%       \\
                            & VTN-SlowFast-ViT & 0.67 & 0.65 & 74.24\%     & 49.78\%       \\
                            & MViTv2 \citep{li2022mvitv2} & 0.68 & 0.67 & 73.80\%     & 48.47\%       \\
                            \midrule
\multirow{3}{*}{EPIC-Tent}  & TSN \citep{wang2016temporal} & 0.94 & 0.78 & \textbf{73.47\%}     & 42.86\%       \\
                            & SlowFast-R50 \citep{feichtenhofer2019slowfast}  & \textbf{0.79} & \textbf{0.72} & 72.79\%     & \textbf{45.58\%}       \\
                            & VTN-SlowFast-ViT & 0.85 & 0.76 & 68.71\%     & 39.46\%       \\
                            & MViTv2 \citep{li2022mvitv2} & 0.94 & 0.81 & 62.58\%     & 37.42\%       \\
                            \midrule
\multirow{3}{*}{EPIC-Tower} & TSN \citep{wang2016temporal} & 0.93 & 0.76 & 85.19\%     & 57.41\%       \\
                            & SlowFast-R50 \citep{feichtenhofer2019slowfast} & 0.78 & \textbf{0.69} & 83.33\%     & 51.85\%       \\
                            & VTN-SlowFast-ViT & 0.88 & 0.72 & 85.19\%     & \textbf{59.26\%}        \\
                            & MViTv2 \citep{li2022mvitv2} & \textbf{0.76} & 0.73 & \textbf{88.89\%}     & 37.04\%       \\
                            \bottomrule
\end{tabular}
\end{table*}

\begin{table*}[ht]
\scriptsize
\centering
\caption{\textbf{Experiment results for struggle regression---split 3.}}
\label{tab:reg split 3}
\begin{tabular}{c|l|cccc}
\toprule
\multirow{2}{*}{Activities} & \multirow{2}{*}{Models} & \multicolumn{4}{c}{Regression \&   Classification - Split 3} \\
                            &                  & MSE$\downarrow$  & MAE$\downarrow$  & Binary Acc.$\uparrow$ & Four-way Acc.$\uparrow$ \\
                            \midrule
\multirow{3}{*}{EPIC-Pipes} & TSN \citep{wang2016temporal}  & 0.83 & 0.74 & 71.31\%     & 42.21\%       \\
                            & SlowFast-R50 \citep{feichtenhofer2019slowfast} & 0.61 & 0.63 & 78.69\%     & 47.95\%       \\
                            & VTN-SlowFast-ViT & \textbf{0.56} & \textbf{0.61} & \textbf{80.33\%}     & \textbf{51.23\%}       \\
                            & MViTv2 \citep{li2022mvitv2} & 0.73 & 0.68 & 75.00\%     & 47.54\%       \\
                            \midrule
\multirow{3}{*}{EPIC-Tent}  & TSN \citep{wang2016temporal}  & 0.87 & 0.76 & 65.75\%     & \textbf{43.84\%}       \\
                            & SlowFast-R50 \citep{feichtenhofer2019slowfast} & \textbf{0.82} & \textbf{0.75} & \textbf{68.49\%}     & \textbf{43.84\%}       \\
                            & VTN-SlowFast-ViT & 0.87 & 0.77 & 63.70\%     & 41.10\%       \\
                            & MViTv2 \citep{li2022mvitv2} & 0.96 & 0.81 & 56.16\%     & 36.99\%       \\
                            \midrule
\multirow{3}{*}{EPIC-Tower} & TSN \citep{wang2016temporal} & 0.87 & 0.71 & 88.06\%     & 64.18\%       \\
                            & SlowFast-R50 \citep{feichtenhofer2019slowfast} & \textbf{0.49} & \textbf{0.54} & \textbf{92.54\%}     & 65.67\%       \\
                            & VTN-SlowFast-ViT & 0.58 & 0.62 & \textbf{92.54\%}     & \textbf{71.64\%}         \\
                            & MViTv2 \citep{li2022mvitv2} & 1.19 & 0.91 & 76.12\%     & 35.82\%       \\
                            \bottomrule
\end{tabular}
\end{table*}

\begin{table*}[ht]
\scriptsize
\centering
\caption{\textbf{Experiment results for struggle regression---split 4.}}
\label{tab:reg split 4}
\begin{tabular}{c|l|cccc}
\toprule
\multirow{2}{*}{Activities} & \multirow{2}{*}{Models} & \multicolumn{4}{c}{Regression \&   Classification - Split 4} \\
                            &                  & MSE$\downarrow$  & MAE$\downarrow$  & Binary Acc.$\uparrow$ & Four-way Acc.$\uparrow$ \\
                            \midrule
\multirow{3}{*}{EPIC-Pipes} & TSN \citep{wang2016temporal} & 0.89 & 0.79 & 69.58\%   & 41.83\%   \\
                            & SlowFast-R50 \citep{feichtenhofer2019slowfast}    & 0.73 & 0.70  & \textbf{77.57\%}   & 41.44\%    \\
                            & VTN-SlowFast-ViT & \textbf{0.68} & \textbf{0.68} & \textbf{77.19\%}     & \textbf{44.87\%}       \\
                            & MViTv2 \citep{li2022mvitv2} & 0.82 & 0.75 & 76.05\%     & 40.30\%       \\
                            \midrule
\multirow{3}{*}{EPIC-Tent}  & TSN \citep{wang2016temporal} & \textbf{0.92} & \textbf{0.79} & \textbf{64.29\%}     & 35.06\%       \\
                            & SlowFast-R50 \citep{feichtenhofer2019slowfast}  & 1.04 & 0.85 & 61.04\%     & 35.06\%       \\
                            & VTN-SlowFast-ViT & 0.97 & 0.81 & \textbf{64.29\%}     & \textbf{40.91\%}       \\
                            & MViTv2 \citep{li2022mvitv2} & 1.33 & 0.96 & 59.09\%     & 32.47\%       \\
                            \midrule
\multirow{3}{*}{EPIC-Tower} & TSN \citep{wang2016temporal}  & 1.19 & 0.84 & 82.00\%   & 52.00\%    \\
                            & SlowFast-R50 \citep{feichtenhofer2019slowfast}  & 1.13 & 0.83 & 84.00\%  & 48.00\%   \\
                            & VTN-SlowFast-ViT & \textbf{0.73} & \textbf{0.68} & \textbf{86.00\%}     & \textbf{56.00\%}         \\
                            & MViTv2 \citep{li2022mvitv2} & 1.05 & 0.89 & 74.00\%     & 30.00\%       \\
                            \bottomrule
\end{tabular}
\end{table*}

\begin{table*}[ht]
\centering
\caption{\textbf{Experiment results for struggle regression---results across all four splits including mean values and standard deviation (StdDev).}}
\label{tab:reg stats mean-std}
\resizebox{\textwidth}{!}{%
\begin{tabular}{c|l|cccc}
\toprule
\multirow{2}{*}{Activities} & \multirow{2}{*}{Models} & \multicolumn{4}{c}{Regression \&   Classification - Statistics (Mean\textpm StdDev)} \\ \cmidrule{3-6} 
                            &                  & MSE$\downarrow$   & MAE$\downarrow$   & Binary Acc.$\uparrow$ & Four-way Acc.$\uparrow$ \\
                            \midrule
\multirow{3}{*}{EPIC-Pipes} & TSN \citep{wang2016temporal}  & 0.798\textpm0.094 & 0.725\textpm0.050 & 72.42\textpm2.52\%     & 45.02\textpm3.38\% \\
                            & SlowFast-R50 \citep{feichtenhofer2019slowfast}    & \textbf{0.645\textpm0.051} & 0.655\textpm0.032 & 76.89\textpm2.02\%     & 45.28\textpm3.00\% \\
                            & VTN-SlowFast-ViT & 0.648\textpm0.051 & \textbf{0.650\textpm0.025}  & \textbf{77.30\textpm2.16\%}   & \textbf{48.56\textpm2.36\%}  \\
                            & MViTv2 \citep{li2022mvitv2} & 0.745\textpm0.058 & 0.708\textpm0.039  & 75.03\textpm0.93\%   & 44.71\textpm3.93\%  \\
                            \midrule
\multirow{3}{*}{EPIC-Tent}  & TSN \citep{wang2016temporal}  & 0.888\textpm0.047  & 0.768\textpm0.019 & 68.27\textpm3.57\%  & 40.95\textpm3.46\%   \\
                            & SlowFast-R50 \citep{feichtenhofer2019slowfast}    & \textbf{0.850\textpm0.112}  & \textbf{0.760\textpm0.053}  & \textbf{69.06\textpm5.05\%}     & \textbf{41.63\textpm3.99\%}   \\
                            & VTN-SlowFast-ViT & 0.868\textpm0.068 & 0.770\textpm0.025  & 66.39\textpm2.40\%     & 40.69\textpm0.72\%  \\
                            & MViTv2 \citep{li2022mvitv2} & 1.055\textpm0.184 & 0.858\textpm0.071  & 59.86\textpm2.87\%   & 35.60\textpm2.24\%  \\
                            \midrule
\multirow{3}{*}{EPIC-Tower} & TSN \citep{wang2016temporal}  & 0.935\textpm0.161 & 0.748\textpm0.061 & 85.35\textpm2.19\%  & 58.01\textpm4.32\%   \\
                            & SlowFast-R50 \citep{feichtenhofer2019slowfast}  & 0.793\textpm0.227 & 0.690\textpm0.103  & 86.89\textpm3.66\%     & 56.00\textpm6.72\%   \\
                            & VTN-SlowFast-ViT & \textbf{0.723\textpm0.107} & \textbf{0.673\textpm0.036} & \textbf{87.86\textpm2.85\%}     & \textbf{61.73\textpm5.92\%}  \\
                            & MViTv2 \citep{li2022mvitv2} & 0.955\textpm0.200 & 0.828\textpm0.087  & 81.29\textpm7.33\%   & 33.41\textpm3.54\%  \\
                            \bottomrule
\end{tabular}%
}
\end{table*} 

\begin{table*}[ht]
\centering
\caption{\textbf{Struggle label distribution learning experiment results on split 1.}}
\label{tab:ldl split 1}
\resizebox{\textwidth}{!}{
\begin{tabular}{c|l|cccc}
\toprule
\multirow{2}{*}{Activities} & \multirow{2}{*}{Models} & \multicolumn{4}{c}{Label Distribution   Learning - Split 1} \\
                            &                  & MAE$\downarrow$  & Spearman's Rho$\uparrow$ & Binary Cls.$\uparrow$ & Four-way Cls.$\uparrow$ \\
                            \midrule
\multirow{3}{*}{EPIC-Pipes} & TSN \citep{wang2016temporal} & 0.12 & 0.4922         & \textbf{80.00\%}     & 39.27\%       \\
                            & SlowFast-R50 \citep{feichtenhofer2019slowfast}  & \textbf{0.11} & \textbf{0.5484}    & \textbf{80.00\%}  & \textbf{44.73\%}   \\
                            & VTN-SlowFast-ViT & \textbf{0.11} & 0.5316         & 78.55\%     & 38.55\%       \\
                            & MViTv2 \citep{li2022mvitv2} & 0.12 & 0.4876         & 78.18\%     & 39.64\%       \\
                            \midrule
\multirow{3}{*}{EPIC-Tent}  & TSN \citep{wang2016temporal} & 0.16 & 0.2743         & 52.17\%     & 29.71\%     \\
                            & SlowFast-R50 \citep{feichtenhofer2019slowfast} & \textbf{0.13} & 0.3658   & \textbf{71.74\%}  & 37.68\%   \\
                            & VTN-SlowFast-ViT & \textbf{0.13} & \textbf{0.3785}         & 64.49\%     & 36.96\%       \\
                            & MViTv2 \citep{li2022mvitv2} & \textbf{0.13} & 0.3302         & 69.56\%     & \textbf{40.58\%}       \\
                            \midrule
\multirow{3}{*}{EPIC-Tower} & TSN \citep{wang2016temporal}  & 0.30  & 0.4881         & 73.85\%     & 35.38\%       \\
                            & SlowFast-R50 \citep{feichtenhofer2019slowfast} & 0.14 & 0.4938    & 73.85\%  & 30.77\%   \\
                            & VTN-SlowFast-ViT & 0.13 & \textbf{0.5709}         & \textbf{76.92\%}     & 44.62\%       \\
                            & MViTv2 \citep{li2022mvitv2} & \textbf{0.12} & 0.5535         & \textbf{76.92\%}     & \textbf{47.69\%}       \\
                            \bottomrule
\end{tabular}
}
\end{table*}

\begin{table*}[ht]
\centering
\caption{\textbf{Struggle label distribution learning experiment results on split 2.}}
\label{tab:ldl split 2}
\resizebox{\textwidth}{!}{
\begin{tabular}{c|l|cccc}
\toprule
\multirow{2}{*}{Activities} & \multirow{2}{*}{Models} & \multicolumn{4}{c}{Label Distribution   Learning - Split 2} \\
                            &                  & MAE$\downarrow$  & Spearman's Rho$\uparrow$ & Binary Cls.$\uparrow$ & Four-way Cls.$\uparrow$ \\
                            \midrule
\multirow{3}{*}{EPIC-Pipes} & TSN \citep{wang2016temporal} & 0.13 & 0.4771         & \textbf{78.17\%}     & 41.92\%       \\
                            & SlowFast-R50 \citep{feichtenhofer2019slowfast}   & \textbf{0.12} & 0.5012   & 77.73\%  & \textbf{45.41\%}   \\
                            & VTN-SlowFast-ViT & \textbf{0.12} & \textbf{0.5113}         & \textbf{78.17\%}     & 42.79\%       \\
                            & MViTv2 \citep{li2022mvitv2} & \textbf{0.12} & 0.5047         & 77.73\%     & 44.98\%       \\
                            \midrule
\multirow{3}{*}{EPIC-Tent}  & TSN \citep{wang2016temporal}  & 0.18 & 0.1983    & 50.34\%   & 21.09\%    \\
                            & SlowFast-R50 \citep{feichtenhofer2019slowfast}   & 0.14 & 0.2048    & 53.74\%    & 27.21\%   \\
                            & VTN-SlowFast-ViT & \textbf{0.13} & \textbf{0.2941}         & \textbf{63.95\%}     & \textbf{38.10\%}       \\
                            & MViTv2 \citep{li2022mvitv2} & 0.14 & 0.2294         & 61.22\%     & 34.69\%       \\
                            \midrule
\multirow{3}{*}{EPIC-Tower} & TSN \citep{wang2016temporal}  & 0.17 & 0.4137   & 72.22\%    & 37.04\%     \\
                            & SlowFast-R50 \citep{feichtenhofer2019slowfast}  & 0.14 & 0.5532    & 75.93\%   & 48.15\%    \\
                            & VTN-SlowFast-ViT & \textbf{0.12} & 0.5879         & 77.78\%     & 57.41\%       \\
                            & MViTv2 \citep{li2022mvitv2} & \textbf{0.12} & \textbf{0.6738}         & \textbf{87.04\%}     & \textbf{59.26\%}       \\
                            \bottomrule
\end{tabular}
}
\end{table*}

\begin{table*}[ht]
\centering
\caption{\textbf{Struggle label distribution learning experiment results on split 3.}}
\label{tab:ldl split 3}
\resizebox{\textwidth}{!}{
\begin{tabular}{c|l|cccc}
\toprule
\multirow{2}{*}{Activities} & \multirow{2}{*}{Models} & \multicolumn{4}{c}{Label Distribution   Learning - Split 3} \\
                            &                  & MAE$\downarrow$  & Spearman's Rho$\uparrow$ & Binary Cls.$\uparrow$ & Four-way Cls.$\uparrow$ \\
                            \midrule
\multirow{3}{*}{EPIC-Pipes} & TSN \citep{wang2016temporal}  & \textbf{0.12} & 0.5605         & \textbf{82.38\%}     & \textbf{55.33\%}       \\
                            & SlowFast-R50 \citep{feichtenhofer2019slowfast}  & 0.13 & 0.5940    & 79.51\%   & 53.69\%   \\
                            & VTN-SlowFast-ViT & \textbf{0.12} & \textbf{0.6202}         & 81.97\%     & 52.05\%       \\
                            & MViTv2 \citep{li2022mvitv2} & 0.13 & 0.5533         & 77.46\%     & 54.92\%       \\
                            \midrule
\multirow{3}{*}{EPIC-Tent}  & TSN \citep{wang2016temporal}  & 0.17 & 0.2733         & 63.01\%     & 34.93\%       \\
                            & SlowFast-R50 \citep{feichtenhofer2019slowfast}  & \textbf{0.13} & 0.3282     & 54.11\%   & 32.19\%   \\
                            & VTN-SlowFast-ViT & \textbf{0.13} & \textbf{0.3536}         & 63.01\%     & 33.56\%       \\
                            & MViTv2 \citep{li2022mvitv2} & \textbf{0.13} & 0.3449         & \textbf{67.81\%}     & \textbf{36.99\%}       \\
                            \midrule
\multirow{3}{*}{EPIC-Tower} & TSN \citep{wang2016temporal} & 0.24 & 0.4273         & 70.15\%     & 35.82\%       \\
                            & SlowFast-R50 \citep{feichtenhofer2019slowfast}  & 0.15 & 0.4739     & 71.64\%   & 32.84\%    \\
                            & VTN-SlowFast-ViT & \textbf{0.13} & \textbf{0.4985}         & \textbf{77.61\%}     & \textbf{44.78\%}       \\
                            & MViTv2 \citep{li2022mvitv2} & 0.16 & 0.3775         & 73.13\%     & 40.30\%       \\
                            \bottomrule
\end{tabular}
}
\end{table*}

\begin{table*}[ht]
\centering
\caption{\textbf{Struggle label distribution learning experiment results on split 4.}}
\label{tab:ldl split 4}
\resizebox{\textwidth}{!}{
\begin{tabular}{c|l|cccc}
\toprule
\multirow{2}{*}{Activities} & \multirow{2}{*}{Models} & \multicolumn{4}{c}{Label Distribution   Learning - Split 4} \\
                            &                  & MAE$\downarrow$  & Spearman's Rho$\uparrow$ & Binary Cls.$\uparrow$ & Four-way Cls.$\uparrow$ \\
                            \midrule
\multirow{3}{*}{EPIC-Pipes} & TSN \citep{wang2016temporal} & 0.13 & 0.5181         & 78.33\%     & 47.53\%       \\
                            & SlowFast-R50 \citep{feichtenhofer2019slowfast}  & \textbf{0.12} & 0.6020          & \textbf{82.13\%}     & 51.33\%       \\
                            & VTN-SlowFast-ViT & \textbf{0.12} & \textbf{0.6125}         & \textbf{82.51\%}     & \textbf{52.47\%}       \\
                            & MViTv2 \citep{li2022mvitv2} & \textbf{0.12} & 0.5866         & \textbf{82.13\%}     & 51.33\%       \\
                            \midrule
\multirow{3}{*}{EPIC-Tent}  & TSN \citep{wang2016temporal}  & 0.17 & 0.1769         & 50.65\%     & 27.92\%       \\
                            & SlowFast-R50 \citep{feichtenhofer2019slowfast}  & \textbf{0.14} & 0.2560      & 48.70\%   & 27.27\%   \\
                            & VTN-SlowFast-ViT & \textbf{0.14} & \textbf{0.3011}         & \textbf{61.69\%}     & 31.17\%       \\
                            & MViTv2 \citep{li2022mvitv2} & \textbf{0.14} & 0.1873         & 59.74\%     & \textbf{36.36\%}       \\
                            \midrule
\multirow{3}{*}{EPIC-Tower} & TSN \citep{wang2016temporal}  & 0.24 & 0.3382         & 68.00\%     & 40.00\%       \\
                            & SlowFast-R50 \citep{feichtenhofer2019slowfast}  & 0.15 & 0.4534     & 78.00\%   & 50.00\%   \\
                            & VTN-SlowFast-ViT & \textbf{0.12} & \textbf{0.6183}         & 78.00\%     & 54.00\%       \\
                            & MViTv2 \citep{li2022mvitv2} & 0.13 & 0.5008         & \textbf{84.00\%}     & \textbf{56.00\%}       \\
                            \bottomrule
\end{tabular}
}
\end{table*}

\begin{table*}[ht]
\centering
\caption{\textbf{Struggle label distribution learning experiment results---results across all the four splits including mean values and standard deviation (StdDev).}}
\label{tab:ldl stats mean stddev}
\resizebox{\textwidth}{!}{%
\begin{tabular}{c|l|cccc}
\toprule
\multirow{2}{*}{Activities} &
  \multirow{2}{*}{Models} &
  \multicolumn{4}{c}{Label Distribution Learning - Statistics (Mean\textpm StdDev)} \\ \cmidrule{3-6} 
 &   &  MAE$\downarrow$ &  Spearman's Rho$\uparrow$ &  Binary Cls.$\uparrow$ &  Four-way Cls.$\uparrow$ \\ \midrule
\multirow{3}{*}{EPIC-Pipes} & TSN \citep{wang2016temporal} & 0.13\textpm0.01 & 0.5120\textpm0.0365  & 79.72\textpm1.96\% & 46.01\textpm7.10\%  \\
                            & SlowFast-R50 \citep{feichtenhofer2019slowfast}  & \textbf{0.12\textpm0.01} & 0.5614\textpm0.0466 & 79.84\textpm1.81\% & \textbf{48.79\textpm4.41\%}  \\
                            & VTN-SlowFast-ViT & \textbf{0.12\textpm0.01} & \textbf{0.5689\textpm0.0555} & \textbf{80.30\textpm2.26\%} & 46.47\textpm6.91\%  \\
                            & MViTv2 \citep{li2022mvitv2} & \textbf{0.12\textpm0.00} & 0.5331\textpm0.0453 & 78.88\textpm2.19\% & 47.72\textpm6.77\%  \\
                            \midrule
\multirow{3}{*}{EPIC-Tent}  & TSN \citep{wang2016temporal} & 0.17\textpm0.01 & 0.2307\textpm0.0505 & 54.04\textpm6.03\% & 28.41\textpm5.72\%   \\
                            & SlowFast-R50 \citep{feichtenhofer2019slowfast}   & 0.14\textpm0.01 & 0.2887\textpm0.0721 & 57.07\textpm10.08\% & 31.09\textpm4.98\%  \\
                            & VTN-SlowFast-ViT & \textbf{0.13\textpm0.01} & \textbf{0.3318\textpm0.0409} & \textbf{63.29\textpm1.23\%} & 34.95\textpm3.17\%  \\
                            & MViTv2 \citep{li2022mvitv2} & \textbf{0.13\textpm0.00} & 0.2730\textpm0.0768 & \textbf{64.58\textpm4.83\%} & \textbf{37.16\textpm2.48\%}  \\
                            \midrule
\multirow{3}{*}{EPIC-Tower} & TSN \citep{wang2016temporal} & 0.24\textpm0.05 & 0.4168\textpm0.0616 & 71.06\textpm2.54\% & 37.06\textpm2.08\%  \\
                            & SlowFast-R50 \citep{feichtenhofer2019slowfast}   & 0.15\textpm0.01 & 0.4936\textpm0.0430 & 74.86\textpm2.73\% & 40.44\textpm10.04\%  \\
                            & VTN-SlowFast-ViT & \textbf{0.13\textpm0.01} & \textbf{0.5689\textpm0.0509} & \textbf{77.58\textpm0.47\%} & \textbf{50.20\textpm6.50\%}  \\
                            & MViTv2 \citep{li2022mvitv2} & 0.14\textpm0.02 & 0.5264\textpm0.1229 & \textbf{80.27\textpm6.38\%} & \textbf{50.81\textpm8.53\%}  \\
                            \bottomrule
\end{tabular}%
}
\end{table*}

\begin{table*}[ht]
\centering
\caption{\textbf{Experiment results using one frame as input}. The original experiment results for the classification task compared with the one-frame experiment results of training and testing the models using only one frame. We report the results by calculating the mean and standard deviation (StdDev) over the four splits. $\Delta$ Mean = Mean(train\&test) - Mean(test only).}
\label{table:one frame}
\centering
\resizebox{\textwidth}{!}{%
\begin{tabular}{c|l|cccccccccc}
\toprule
\multirow{4}{*}{Activities}    & \multirow{4}{*}{Models}     & \multicolumn{10}{c}{Top1 Accuracy Rate (\%)}                                                  \\
                             &                             & \multicolumn{5}{c}{Binary Classification}     & \multicolumn{5}{c}{Four-way Classification}   \\
 &
   &
  \multicolumn{2}{c}{Only Test} &
  \multicolumn{2}{c}{Train \& Test} &
  \multirow{2}{*}{\begin{tabular}[c]{@{}c@{}}$\Delta$ Mean\end{tabular}} &
  \multicolumn{2}{c}{Only Test} &
  \multicolumn{2}{c}{Train \& Test} &
  \multirow{2}{*}{\begin{tabular}[c]{@{}c@{}}$\Delta$ Mean\end{tabular}} \\
                             &                             & StdDev & Mean    & StdDev & Mean    &         & StdDev & Mean    & StdDev & Mean    &         \\ \midrule
\multirow{16}{*}{Pipes-Struggle} & TSN \citep{wang2016temporal} & 2.10\% & 74.54\% & 2.10\% & 74.54\% & -\%     & 3.57\% & 48.59\% & 3.57\% & 48.59\% & -\%     \\
                             & One-frame test 25\%         & 2.99\% & 64.60\% & 2.65\% & 67.18\% & 2.58\%  & 3.87\% & 38.55\% & 3.00\% & 42.10\% & 3.55\%  \\
                             & One-frame test 50\%         & 2.80\% & 67.10\% & 3.35\% & 69.07\% & 1.97\%  & 2.95\% & 40.85\% & 3.57\% & 42.31\% & 1.46\%  \\
                             & One-frame test 75\%         & 3.79\% & 67.39\% & 4.63\% & 70.19\% & 2.81\%  & 3.49\% & 41.73\% & 3.58\% & 42.33\% & 0.61\%  \\ \cmidrule{2-12}
                             & SlowFast-R50 \citep{feichtenhofer2019slowfast}  & 2.44\% & 78.01\% & 2.44\% & 78.01\% & -\%     & 1.97\% & 50.77\% & 1.97\% & 50.77\% & -\%     \\
                             & One-frame test 25\%         & 5.38\% & 63.24\% & 3.03\% & 63.25\% & 0.01\%  & 3.90\% & 38.58\% & 1.85\% & 39.51\% & 0.94\%  \\
                             & One-frame test 50\%         & 4.39\% & 62.25\% & 3.11\% & 63.66\% & 1.41\%  & 3.23\% & 38.67\% & 2.02\% & 39.79\% & 1.11\%  \\
                             & One-frame test 75\%         & 7.45\% & 62.31\% & 2.95\% & 63.36\% & 1.05\%  & 3.55\% & 40.36\% & 1.89\% & 39.60\% & -0.75\% \\ \cmidrule{2-12}
                             & SlowFast-gMLP  & 1.62\% & 78.59\% & 1.62\% & 78.59\% & -\%     & 2.97\% & 49.70\% & 2.97\% & 49.70\% & -\%     \\
                             & One-frame test 25\%         & 3.72\% & 62.60\% & 3.24\% & 64.92\% & 2.32\%  & 6.99\% & 31.34\% & 2.14\% & 38.84\% & 7.50\%  \\
                             & One-frame test 50\%         & 3.72\% & 62.60\% & 3.68\% & 67.74\% & 5.14\%  & 7.46\% & 31.73\% & 2.69\% & 41.87\% & 10.15\% \\
                             & One-frame test 75\%         & 3.72\% & 62.60\% & 2.40\% & 65.24\% & 2.64\%  & 6.40\% & 29.62\% & 1.99\% & 40.74\% & 11.12\% \\ \cmidrule{2-12}
                             & VTN-SlowFast-ViT  & 1.66\% & 77.80\% & 1.66\% & 77.80\% & -\%     & 1.58\% & 50.83\% & 1.58\% & 50.83\% & -\%     \\
                             & One-frame test 25\%         & 3.83\% & 62.89\% & 3.27\% & 63.98\% & 1.09\%  & 4.26\% & 34.10\% & 1.97\% & 39.82\% & 5.72\%  \\
                             & One-frame test 50\%         & 3.73\% & 62.79\% & 3.04\% & 63.77\% & 0.98\%  & 4.58\% & 35.36\% & 2.42\% & 39.69\% & 4.33\%  \\
                             & One-frame test 75\%         & 3.73\% & 62.21\% & 2.75\% & 63.66\% & 1.45\%  & 4.87\% & 34.14\% & 1.84\% & 40.24\% & 6.10\%  \\ 
                             \midrule
\multirow{16}{*}{Tent-Struggle}  & TSN \citep{wang2016temporal}   & 5.75\% & 68.05\% & 5.75\% & 68.05\% & -\%     & 3.87\% & 38.23\% & 3.87\% & 38.23\% & -\%     \\
                             & One-frame test 25\%         & 6.09\% & 58.49\% & 3.06\% & 59.58\% & 1.10\%  & 5.46\% & 31.77\% & 4.48\% & 37.86\% & 6.09\%  \\
                             & One-frame test 50\%         & 3.74\% & 61.88\% & 6.34\% & 61.07\% & -0.81\% & 5.07\% & 31.66\% & 4.38\% & 36.81\% & 5.15\%  \\
                             & One-frame test 75\%         & 3.33\% & 58.72\% & 5.32\% & 61.84\% & 3.12\%  & 5.50\% & 33.59\% & 4.07\% & 35.79\% & 2.20\%  \\ \cmidrule{2-12}
                             & SlowFast-R50 \citep{feichtenhofer2019slowfast}  & 5.98\% & 69.56\% & 5.98\% & 69.56\% & 0.00\%  & 4.68\% & 39.48\% & 4.68\% & 39.48\% & -\%     \\
                             & One-frame test 25\%         & 3.08\% & 55.82\% & 2.64\% & 55.24\% & -0.57\% & 3.28\% & 29.67\% & 4.73\% & 38.02\% & 8.35\%  \\
                             & One-frame test 50\%         & 2.69\% & 55.79\% & 4.03\% & 58.22\% & 2.43\%  & 5.64\% & 29.20\% & 3.82\% & 36.32\% & 7.12\%  \\
                             & One-frame test 75\%         & 2.43\% & 55.07\% & 4.00\% & 56.82\% & 1.75\%  & 4.24\% & 29.33\% & 4.59\% & 35.27\% & 5.93\%  \\ \cmidrule{2-12}
                             & SlowFast-gMLP  & 6.16\% & 68.00\% & 6.16\% & 68.00\% & -\%     & 1.96\% & 39.52\% & 1.96\% & 39.52\% & -\%     \\
                             & One-frame test 25\%         & 3.37\% & 52.38\% & 2.99\% & 55.64\% & 3.26\%  & 8.21\% & 31.66\% & 3.07\% & 39.04\% & 7.38\%  \\
                             & One-frame test 50\%         & 7.84\% & 52.97\% & 4.63\% & 57.53\% & 4.57\%  & 5.59\% & 33.11\% & 4.21\% & 34.48\% & 1.37\%  \\
                             & One-frame test 75\%         & 2.99\% & 53.36\% & 5.69\% & 52.94\% & -0.42\% & 5.87\% & 32.47\% & 2.89\% & 33.00\% & 0.53\%  \\ \cmidrule{2-12}
                             & VTN-SlowFast-ViT  & 5.70\% & 68.53\% & 5.70\% & 68.53\% & -\%     & 3.56\% & 40.11\% & 3.56\% & 40.11\% & -\%     \\
                             & One-frame test 25\%         & 2.84\% & 54.76\% & 3.73\% & 56.53\% & 1.77\%  & 6.31\% & 33.06\% & 4.51\% & 37.16\% & 4.10\%  \\
                             & One-frame test 50\%         & 3.03\% & 55.84\% & 2.71\% & 55.27\% & -0.56\% & 7.26\% & 33.45\% & 4.07\% & 35.47\% & 2.02\%  \\
                             & One-frame test 75\%         & 3.07\% & 55.13\% & 3.50\% & 54.34\% & -0.79\% & 3.32\% & 35.69\% & 3.79\% & 35.95\% & 0.25\%  \\
                             \midrule
\multirow{16}{*}{Tower-Struggle} & TSN \citep{wang2016temporal} & 4.41\% & 82.59\% & 4.41\% & 82.59\% & -\%     & 6.72\% & 59.49\% & 6.72\% & 59.49\% & -\%     \\
                             & One-frame test 25\%         & 6.40\% & 77.05\% & 3.39\% & 79.74\% & 2.69\%  & 4.85\% & 54.50\% & 4.73\% & 57.56\% & 3.07\%  \\
                             & One-frame test 50\%         & 5.85\% & 73.66\% & 3.39\% & 79.74\% & 6.07\%  & 7.02\% & 51.95\% & 4.73\% & 57.56\% & 5.61\%  \\
                             & One-frame test 75\%         & 3.65\% & 72.64\% & 3.39\% & 79.74\% & 7.10\%  & 6.31\% & 51.50\% & 4.73\% & 57.56\% & 6.07\%  \\ \cmidrule{2-12}
                             & SlowFast-R50 \citep{feichtenhofer2019slowfast}  & 3.36\% & 88.24\% & 3.36\% & 88.24\% & -\%     & 1.91\% & 63.92\% & 1.91\% & 63.92\% & -\%     \\
                             & One-frame test 25\%         & 3.45\% & 77.61\% & 4.28\% & 81.89\% & 4.28\%  & 6.29\% & 55.34\% & 4.73\% & 57.56\% & 2.23\%  \\
                             & One-frame test 50\%         & 4.29\% & 80.48\% & 4.21\% & 81.98\% & 1.49\%  & 4.43\% & 55.54\% & 4.73\% & 57.56\% & 2.02\%  \\
                             & One-frame test 75\%         & 3.01\% & 78.99\% & 3.25\% & 80.74\% & 1.75\%  & 2.73\% & 55.78\% & 4.73\% & 57.56\% & 1.79\%  \\ \cmidrule{2-12}
                             & SlowFast-gMLP   & 2.75\% & 87.86\% & 2.75\% & 87.86\% & -\%     & 2.57\% & 66.40\% & 2.57\% & 66.40\% & -\%     \\
                             & One-frame test 25\%         & 3.39\% & 79.74\% & 3.39\% & 79.74\% & 0.00\%  & 4.73\% & 57.56\% & 4.60\% & 58.44\% & 0.87\%  \\
                             & One-frame test 50\%         & 3.39\% & 79.74\% & 3.39\% & 79.74\% & 0.00\%  & 4.73\% & 57.56\% & 4.00\% & 58.06\% & 0.50\%  \\
                             & One-frame test 75\%         & 3.39\% & 79.74\% & 3.39\% & 79.74\% & 0.00\%  & 4.73\% & 57.56\% & 5.23\% & 58.81\% & 1.25\%  \\ \cmidrule{2-12}
                             & VTN-SlowFast-ViT  & 4.54\% & 88.11\% & 4.54\% & 88.11\% & -\%     & 4.09\% & 64.03\% & 4.09\% & 64.03\% & -\%     \\
                             & One-frame test 25\%         & 2.94\% & 79.35\% & 3.39\% & 79.74\% & 0.38\%  & 4.73\% & 57.56\% & 4.73\% & 57.56\% & 0.00\%  \\
                             & One-frame test 50\%         & 2.94\% & 79.35\% & 3.39\% & 79.74\% & 0.38\%  & 4.73\% & 57.56\% & 4.73\% & 57.56\% & 0.00\%  \\
                             & One-frame test 75\%         & 3.39\% & 79.74\% & 3.39\% & 79.74\% & 0.00\%  & 4.73\% & 57.56\% & 4.73\% & 57.56\% & 0.00\% \\
                             \bottomrule
\end{tabular}%
}
\end{table*}

\begin{sidewaystable*}[ht]
\centering
\caption{\textbf{Experiment results using shuffled frames}. Mean Absolute Error (MAE) and Spearman's Rank Correlation (Spearman's Rho) with the corresponding converted classification accuracy for binary classification and four-way classification.}
\label{table:shuffle frame}
\centering
\resizebox{\textwidth}{!}{%
\begin{tabular}{c|l|cccccccccccc}
\toprule
\multirow{3}{*}{Activities} &
  \multirow{3}{*}{Models} &
  \multicolumn{12}{c}{Top1   Accuracy Rate (\%)} \\
 &
   &
  \multicolumn{6}{c}{Binary   Classification} &
  \multicolumn{6}{c}{Four-way   Classification} \\
 &
   &
  Split   1 &
  Split   2 &
  Split   3 &
  Split   4 &
  StdDev &
  Mean &
  Split   1 &
  Split   2 &
  Split   3 &
  Split   4 &
  StdDev &
  Mean \\ \midrule
\multirow{10}{*}{Pipes-Struggle} &
  SlowFast-R50 \citep{feichtenhofer2019slowfast}&
  77.09\% &
  75.11\% &
  80.74\% &
  79.09\% &
  2.44\% &
  78.01\% &
  53.46\% &
  50.66\% &
  48.77\% &
  50.19\% &
  1.97\% &
  50.77\% \\
 &
  Shuffle frames test &
  64.22\% &
  46.46\% &
  60.25\% &
  47.53\% &
  8.96\% &
  54.62\% &
  24.36\% &
  27.60\% &
  30.08\% &
  28.52\% &
  2.41\% &
  27.64\% \\
 &
  Shuffle   Train\&Test &
  75.64\% &
  74.67\% &
  76.23\% &
  78.33\% &
  1.55\% &
  76.22\% &
  48.36\% &
  42.36\% &
  42.62\% &
  49.05\% &
  3.60\% &
  45.60\% \\ \cmidrule{2-14}
 &
  SlowFast-gMLP &
  78.91\% &
  76.42\% &
  80.33\% &
  78.71\% &
  1.62\% &
  78.59\% &
  53.46\% &
  48.04\% &
  46.72\% &
  50.57\% &
  2.97\% &
  49.70\% \\
 &
  Shuffle   frames test &
  41.24\% &
  45.15\% &
  62.62\% &
  47.30\% &
  9.37\% &
  49.08\% &
  28.51\% &
  22.97\% &
  29.67\% &
  22.66\% &
  3.66\% &
  25.95\% \\
 &
  Shuffle   Train\&Test &
  74.91\% &
  74.24\% &
  75.92\% &
  74.90\% &
  0.69\% &
  74.99\% &
  47.27\% &
  44.98\% &
  46.72\% &
  50.95\% &
  2.51\% &
  47.48\% \\ \cmidrule{2-14}
 &
  VTN-SlowFast-ViT &
  77.82\% &
  75.55\% &
  79.51\% &
  78.33\% &
  1.66\% &
  77.80\% &
  50.54\% &
  51.53\% &
  48.77\% &
  52.47\% &
  1.58\% &
  50.83\% \\
 &
  Shuffle   frames test &
  45.09\% &
  53.97\% &
  63.11\% &
  43.42\% &
  9.08\% &
  51.40\% &
  23.27\% &
  32.40\% &
  32.30\% &
  24.18\% &
  4.99\% &
  28.04\% \\
 &
  Shuffle   Train\&Test &
  73.46\% &
  70.74\% &
  73.77\% &
  78.33\% &
  3.15\% &
  74.08\% &
  46.18\% &
  33.19\% &
  40.98\% &
  44.49\% &
  5.77\% &
  41.21\% \\ \midrule
\multirow{10}{*}{Tent-Struggle} &
  SlowFast-R50 \citep{feichtenhofer2019slowfast}&
  76.81\% &
  72.11\% &
  64.38\% &
  64.94\% &
  5.98\% &
  69.56\% &
  45.65\% &
  38.10\% &
  39.73\% &
  34.42\% &
  4.68\% &
  39.48\% \\ 
 &
  Shuffle   frames test &
  50.00\% &
  42.99\% &
  43.56\% &
  48.70\% &
  3.55\% &
  46.31\% &
  22.32\% &
  19.86\% &
  22.60\% &
  26.23\% &
  2.62\% &
  22.75\% \\
 &
  Shuffle   Train\&Test &
  73.19\% &
  75.51\% &
  65.75\% &
  57.14\% &
  8.29\% &
  67.90\% &
  39.13\% &
  39.46\% &
  36.30\% &
  31.17\% &
  3.84\% &
  36.52\% \\ \cmidrule{2-14}
 &
  SlowFast-gMLP &
  73.19\% &
  73.47\% &
  63.01\% &
  62.34\% &
  6.16\% &
  68.00\% &
  41.30\% &
  40.82\% &
  36.99\% &
  38.96\% &
  1.96\% &
  39.52\% \\
 &
  Shuffle   frames test &
  51.16\% &
  45.03\% &
  45.48\% &
  49.74\% &
  3.06\% &
  47.85\% &
  27.97\% &
  29.25\% &
  21.92\% &
  24.42\% &
  3.34\% &
  25.89\% \\
 &
  Shuffle   Train\&Test &
  68.84\% &
  74.15\% &
  65.07\% &
  61.69\% &
  5.34\% &
  67.44\% &
  36.96\% &
  36.74\% &
  31.51\% &
  33.12\% &
  2.70\% &
  34.58\% \\ \cmidrule{2-14}
 &
  VTN-SlowFast-ViT &
  73.91\% &
  72.79\% &
  65.07\% &
  62.34\% &
  5.70\% &
  68.53\% &
  42.75\% &
  40.14\% &
  42.47\% &
  35.06\% &
  3.56\% &
  40.11\% \\
 &
  Shuffle   frames test &
  51.88\% &
  42.99\% &
  44.52\% &
  48.70\% &
  4.04\% &
  47.02\% &
  28.70\% &
  29.93\% &
  25.89\% &
  24.68\% &
  2.43\% &
  27.30\% \\
 &
  Shuffle   Train\&Test &
  68.84\% &
  72.79\% &
  63.01\% &
  59.09\% &
  6.08\% &
  65.93\% &
  41.30\% &
  38.10\% &
  30.14\% &
  28.57\% &
  6.15\% &
  34.53\% \\ \midrule
\multirow{10}{*}{Tower-Struggle} &
  SlowFast-R50 \citep{feichtenhofer2019slowfast}&
  89.23\% &
  85.18\% &
  92.54\% &
  86.00\% &
  3.36\% &
  88.24\% &
  66.15\% &
  64.82\% &
  62.69\% &
  62.00\% &
  1.91\% &
  63.92\% \\
 &
  Shuffle   frames test &
  32.00\% &
  39.63\% &
  49.25\% &
  72.40\% &
  17.54\% &
  48.32\% &
  21.54\% &
  21.48\% &
  42.99\% &
  27.20\% &
  10.15\% &
  28.30\% \\
 &
  Shuffle   Train\&Test &
  81.54\% &
  77.78\% &
  83.58\% &
  82.00\% &
  2.46\% &
  81.23\% &
  63.08\% &
  50.00\% &
  55.22\% &
  62.00\% &
  6.13\% &
  57.58\% \\ \cmidrule{2-14}
 &
  SlowFast-gMLP &
  89.23\% &
  85.18\% &
  91.04\% &
  86.00\% &
  2.75\% &
  87.86\% &
  64.62\% &
  64.82\% &
  70.15\% &
  66.00\% &
  2.57\% &
  66.40\% \\
 &
  Shuffle   frames test &
  47.08\% &
  42.96\% &
  64.18\% &
  64.00\% &
  11.14\% &
  54.56\% &
  28.92\% &
  22.59\% &
  46.57\% &
  36.40\% &
  10.31\% &
  33.62\% \\
 &
  Shuffle   Train\&Test &
  81.54\% &
  75.93\% &
  82.09\% &
  66.00\% &
  7.46\% &
  76.39\% &
  63.08\% &
  48.15\% &
  65.67\% &
  44.00\% &
  10.75\% &
  55.23\% \\ \cmidrule{2-14}
 &
  VTN-SlowFast-ViT &
  89.23\% &
  85.18\% &
  94.03\% &
  84.00\% &
  4.54\% &
  88.11\% &
  67.69\% &
  59.26\% &
  67.16\% &
  62.00\% &
  4.09\% &
  64.03\% \\
 &
  Shuffle   frames test &
  49.54\% &
  42.22\% &
  62.69\% &
  66.40\% &
  11.29\% &
  55.21\% &
  26.46\% &
  19.63\% &
  40.60\% &
  38.00\% &
  9.85\% &
  31.17\% \\
 &
  Shuffle   Train\&Test &
  87.69\% &
  77.78\% &
  86.57\% &
  84.00\% &
  4.43\% &
  84.01\% &
  60.00\% &
  55.56\% &
  62.69\% &
  58.00\% &
  3.02\% &
  59.06\% \\ \bottomrule
\end{tabular}%
}
\end{sidewaystable*}

\clearpage

\end{appendices}


\newpage
\bibliography{sn-bibliography}

\end{document}